\documentclass[11pt,a4paper]{article}
\usepackage[marginratio=1:1,height=600pt,width=460pt,tmargin=100pt]{geometry}
\usepackage[utf8]{inputenc}
\usepackage[parfill]{parskip}    
\usepackage{url}            
\usepackage{booktabs}       
\usepackage{amsfonts}       
\usepackage{nicefrac}       
\usepackage{microtype}      
\usepackage{graphicx}
\usepackage{amsmath}
\usepackage{amssymb}
\graphicspath{{./pics/}}
\usepackage{subcaption}
\usepackage{amsthm}
\usepackage{amscd}
\usepackage{mathrsfs}
\usepackage{bm}
\usepackage{algorithmic, algorithm}
\usepackage{multirow}
\usepackage{mathtools}
\usepackage{hyperref}
\usepackage{url}
\usepackage{enumitem}
\usepackage{authblk}
\usepackage{wrapfig,lipsum,booktabs}

\allowdisplaybreaks

\DeclareMathOperator\supp{supp}
\newcommand{\R}{\mathbb{R}}
\newcommand{\cs}{\mathcal{S}}
\newcommand{\cst}{\mathcal{ST}}
\newtheorem{thm}{Theorem}
\newtheorem*{thm*}{Theorem}
\newtheorem{lemma}{Lemma}
\newtheorem*{lemma*}{Lemma}
\newtheorem{prop}{Proposition}
\newtheorem{rmk}{Remark}

\begin{document}

\title{Scaling-Translation-Equivariant Networks with Decomposed Convolutional Filters}
\author[1]{Wei Zhu\footnote{Email: zhu@math.umass.edu.}}
\author[2]{Qiang Qiu}
\author[3,4]{Robert Calderbank}
\author[4]{Guillermo Sapiro}
\author[3]{Xiuyuan Cheng}

\affil[1]{Department of Mathematics and Statistics, University of Massachusetts Amherst}
\affil[2]{Department of Electrical and Computer Engineering, Purdue University}
\affil[3]{Department of Mathematics, Duke University}
\affil[4]{Department of Electrical and Computer Engineering, Duke University}
\date{}

\maketitle

\begin{abstract}
Encoding the scale information explicitly into the representation learned by a convolutional neural network (CNN) is beneficial for many computer vision tasks especially when dealing with multiscale inputs. We study, in this paper, a scaling-translation-equivariant ($\cst$-equivariant) CNN with joint convolutions across the space and  the scaling group, which is shown to be both sufficient and necessary to achieve equivariance for the regular representation of the scaling-translation group $\cst$. To reduce the model complexity and computational burden,  we decompose the convolutional filters under two pre-fixed separable bases and truncate the expansion to  low-frequency components. A further benefit of the truncated filter expansion is the improved deformation robustness of the equivariant representation, a property which is theoretically analyzed and empirically verified. Numerical experiments demonstrate that the proposed scaling-translation-equivariant network with decomposed convolutional filters (ScDCFNet)
achieves significantly improved performance in multiscale image classification and better interpretability than regular CNNs at a reduced model size.
\end{abstract}

\section{Introduction}

Convolutional neural networks (CNNs) have achieved great success in machine learning problems such as image classification \cite{Krizhevsky_2012_imagenet}, object detection \cite{Ren_2015_detection}, and semantic segmentation \cite{Long_2015_segmentation, Ronneberger_2015_segmentation}. Compared to fully-connected networks, CNNs through spatial weight sharing have the benefit of being translation-equivariant, i.e., translating the input leads to a translated version of the output. This property is crucial for many vision tasks such as image recognition and segmentation. However, regular CNNs are not equivariant to other important group transformations such as  rescaling and rotation, and it is beneficial in some applications to also encode such group information explicitly into the network representation.

Several network architectures have been designed to achieve (2D) roto-translation-equivariance ($SE(2)$-equivariance) \cite{general_e2,Cheng_2018_rotdcf,hexaconv,Worrall_2017_harmonic,bekkers2017template,Zhou_2017_orn,Marcos_2017_rotation,Weiler_2018_learning}, i.e., roughly speaking, if the input is spatially rotated and translated, the output is transformed accordingly. The feature maps of such networks typically include an extra index for the rotation group $SO(2)$. Building on the idea of group convolutions proposed by \cite{Cohen_2016_group} for discrete symmetry groups, \cite{Cheng_2018_rotdcf}, \cite{Worrall_2017_harmonic}, and \cite{Weiler_2018_learning} constructed $SE(2)$-equivariant CNNs by conducting group convolutions jointly across the space and $SO(2)$ using steerable filters \cite{Freeman_1991_steerable}.

Scaling-translation-equivariant ($\cst$-equivariant) CNNs, on the other hand, are usually studied in a less general setting in the existing literature. In particular, joint convolutions across the space and the scaling group $\cs$ are typically not proposed to achieve equivariance in the  general form \cite{Kim_2014_scale, Marcos_2018_scale, Xu_2014_scale, ghosh2019scale}. This is possibly because of two difficulties one encounters when dealing with the scaling group: First, unlike $SO(2)$, it is an acyclic and unbounded group; second, an extra index in $\cs$ incurs a significant increase in model parameters and computational burden. Moreover, due to changing view angle or numerical discretization, the scaling transformation is rarely perfect in practice. One thus needs to quantify and promote the deformation robustness of the equivariant representation (i.e., is the model still ``approximately'' equivariant if the scaling transformation is ``contaminated'' by a nuisance input deformation), which, to the best of our knowledge, has not been studied in prior works.

The purpose of this paper is to address the aforementioned theoretical and practical issues in the construction of $\cst$-equivariant CNN models. Specifically, our contribution is three-fold:
\begin{enumerate}
\item We propose a general $\cst$-equivariant CNN architecture with a joint convolution over $\R^2$ and $\cs$, which is proved in Section~\ref{sec:stability} to be both sufficient and necessary to achieve equivariance for the regular representation of the group $\cst$.
\item A truncated  decomposition of the convolutional filters  under a pre-fixed separable basis on the two geometric domains ($\R^2$ and $\cs$) is used to reduce the model size and computational cost.
\item We prove the representation stability of the proposed architecture up to equivariant scaling action of the input signal, guaranteeing equivariance is ``approximately'' achieved even if the scaling effect is non-perfect. This theoretical result is crucial for the practical implementation of $\cst$-equivariant CNNs when signals are discrete and confined in finite domains.
\end{enumerate}

Our contribution to the family of group-equivariant CNNs is non-trivial; in particular, the scaling group unlike $SO(2)$ is acyclic and non-compact. This poses challenges both in theory and in practice, so that many previous works on group-equivariant CNNs cannot be directly extended. The concurrent independent research by \cite{worrall2019deep}, \cite{sosnovik2019scale}, and \cite{Bekkers2020B-Spline} also discovered  and implemented joint convolutions to achieve scaling-translation-equivariance in the  general form. However, the approach by \cite{worrall2019deep} is limited to scaling factors at only integer powers of 2, and none of the three analyzes and promotes the deformation robustness of the equivariant representation, especially in the practical setting where signals are discretized and computed only on compact domains. We introduce new algorithm design and mathematical techniques to obtain  general $\cst$-equivariant CNNs  with both computational efficiency and proved representation stability.


\section{Related Work}

\textbf{Mixed-scale CNNs.} Incorporating multiscale information into a CNN representation has been studied in many existing works. The Inception net \cite{Szegedy_2015_inception} and its generalizations \cite{Szegedy_2017_inception3, Szegedy_2016_inception2,Li_2019_selective} stack filters of different sizes in a single layer to address the multiscale salient features. Dilated convolutions \cite{Pelt_2018_mixed, Wang_2018_understanding, Yu_2015_multi, Yu_2017_CVPR}, pyramid architectures \cite{ke2017multigrid,lin2017feature}, and multiscale dense networks \cite{huang2017multi} have also been proposed to take into account the multiscale feature information. Although the effectiveness of such models have been empirically  demonstrated  in various vision tasks, there is still a lack of interpretability of their ability to encode the input scale information.

\noindent\textbf{Group-equivariant CNNs.} Group-equivariant CNNs (G-CNNs) have consistently demonstrated their superior performance over classical CNNs by explicitly encoding group information into the network representation. G-CNN was initially proposed by \cite{Cohen_2016_group} to achieve equivariance over finite discrete symmetry groups. The idea is later generalized in \cite{cohen2018general, kondor, steerable_cnn}, and has been applied mainly to discrete or compact continuous groups such as 2D rotation $SO(2)$ \cite{general_e2,Cheng_2018_rotdcf,hexaconv,Worrall_2017_harmonic,bekkers2017template,Zhou_2017_orn,Marcos_2017_rotation,Weiler_2018_learning} and 3D rotation $SO(3)$ \cite{weiler20183d,worrall2018cubenet,thomas2018tensor,spherical,esteves2018learning,winkels20183d,andrearczyk2019exploring}.
  
  Although $\cst$-equivariant (or invariant) CNNs have also been proposed in the literature \cite{Kim_2014_scale, Marcos_2018_scale, Xu_2014_scale, ghosh2019scale}, they are typically studied in a less general setting. In particular, none of these  works proposed to conduct joint convolutions over $\R^2\times \cs$ as a necessary and sufficient condition to achieve equivariance for the regular representation of the group $\cst$, for which reason they are thus variants of a special case of our proposed architecture where the convolutional filters in $\cs$ are Dirac delta functions (c.f. Remark~\ref{rmk:delta}.) The interscale convolution proposed in the independent concurrent works by \cite{worrall2019deep}, \cite{sosnovik2019scale}, and \cite{Bekkers2020B-Spline} bear the most resemblance to our proposed model. In particular, the filter expansion under B-splines for arbitrary Lie groups proposed by \cite{Bekkers2020B-Spline} is also akin to our truncated filter decomposition under compactly supported separable function bases. However, the approach by \cite{worrall2019deep} is limited to scaling factors at only integer powers of 2, and none of the three concurrent works analyzes and promotes deformation robustness of the equivariant representation, which is important both in theory and in practice because scaling effect is rarely perfect due to signal distortion, discretization, and truncation.

\noindent\textbf{Representation stability to input deformations.} Input deformations typically induce noticeable variabilities within object classes, some of which are  uninformative for the vision tasks. Models that are stable to input deformations are thus favorable in many applications. The scattering transform \cite{Bruna_2013_invariant, Mallat_2010_recursive,Mallat_2012_group} computes translation-invariant representations that are Lipschitz continuous to deformations by cascading predefined wavelet transforms and modulus poolings. A joint convolution over $\R^2\times SO(2)$ is later adopted by \cite{Sifre_2013_rotation} to build roto-translation scattering with stable rotation/translation-invariant representations. These models, however, use pre-fixed wavelet transforms in the networks, and are thus nonadaptive to the data. Invariance and stability of deep convolutional representations have also be studied by \cite{bietti2017invariance,bietti2019group} in the context of convolutional kernel networks \cite{NIPS2016_fc8001f8,NIPS2014_81ca0262}. DCFNet \cite{dcfnet} combines  a pre-fixed filter basis and learnable expansion coefficients in a CNN architecture, achieving both data  adaptivity and  representation stability inherited from the filter regularity. This idea is later extended by \cite{Cheng_2018_rotdcf} to produce $SE(2)$-equivariant representations that are Lipschitz continuous in $L^2$ norm to input deformations modulo a global rotation, i.e., the model stays approximately equivariant even if the input rotation is imperfect. To the best of our knowledge, a theoretical analysis of the deformation robustness of a $\cst$-equivariant CNN has yet been studied, and a direct generalization of the result by \cite{Cheng_2018_rotdcf} is futile because the feature maps of a $\cst$-equivariant CNN is typically not in $L^2$ (c.f. Remark~\ref{rmk:l_inf}.)

\section{$\cst$-Equivariant CNN And Filter Decomposition}
\label{sec:scdcf}

Group-equivariance is the property of a mapping $f: X \to Y$ to commute with the group actions on the domain $X$ and codomain $Y$. More specifically, let $G$ be a group, and $D_g$, $T_g$, respectively,  be group actions on $X$ and $Y$. A function $f: X\to Y$ is said to be $G$-equivariant if
\begin{align}
  \label{eq:def_G_equivariance}
  f(D_gx) = T_g(f(x)), \quad \forall~ g\in G, ~x\in X.
\end{align}
$G$-invariance is thus a special case of $G$-equivariance where $T_g = \text{Id}_Y$. For  learning tasks where the feature $y\in Y$ is known a priori to change equivariantly to a group action $g\in G$ on the input $x\in X$, e.g. image segmentation should be equivariant to translation, it would be beneficial to reduce the hypothesis space to include only $G$-equivariant models. In this paper, we consider mainly the scaling-translation group $\cst  =  \cs\ltimes \R^2\cong \R\times \R^2$, which is the semi-direct product between the the scaling group $\cs$ and the translation group $\R^2$. Given  $g = (\beta, v) \in\cst$ and an input image $x^{(0)}(u,\lambda)$ ($u\in\R^2$ is the spatial position, and $\lambda$ is the unstructured channel index, e.g. RGB channels of a color image), the scaling-translation group action $D_g = D_{\beta,v}$ on $x^{(0)}$ is defined as
\begin{align}
  \label{eq:group-action1}
  D_{\beta,v}x^{(0)}(u,\lambda) \coloneqq x^{(0)}\left(2^{-\beta}(u-v),\lambda\right).
\end{align}
Constructing $\cst$-equivariant CNNs thus amounts to finding an architecture $\mathcal{A}$ such that each trained network $f\in\mathcal{A}$ commutes with the group action $D_{\beta,v}$ on the input and a similarly defined group action $T_{\beta,v}$ (to be explained in Section~\ref{sec:secnn}) on the output.

\subsection{$\cst$-Equivariant CNNs}
\label{sec:secnn}

\begin{figure}
  \centering
      \begin{subfigure}[t]{0.49\textwidth}
        \includegraphics[width=\textwidth]{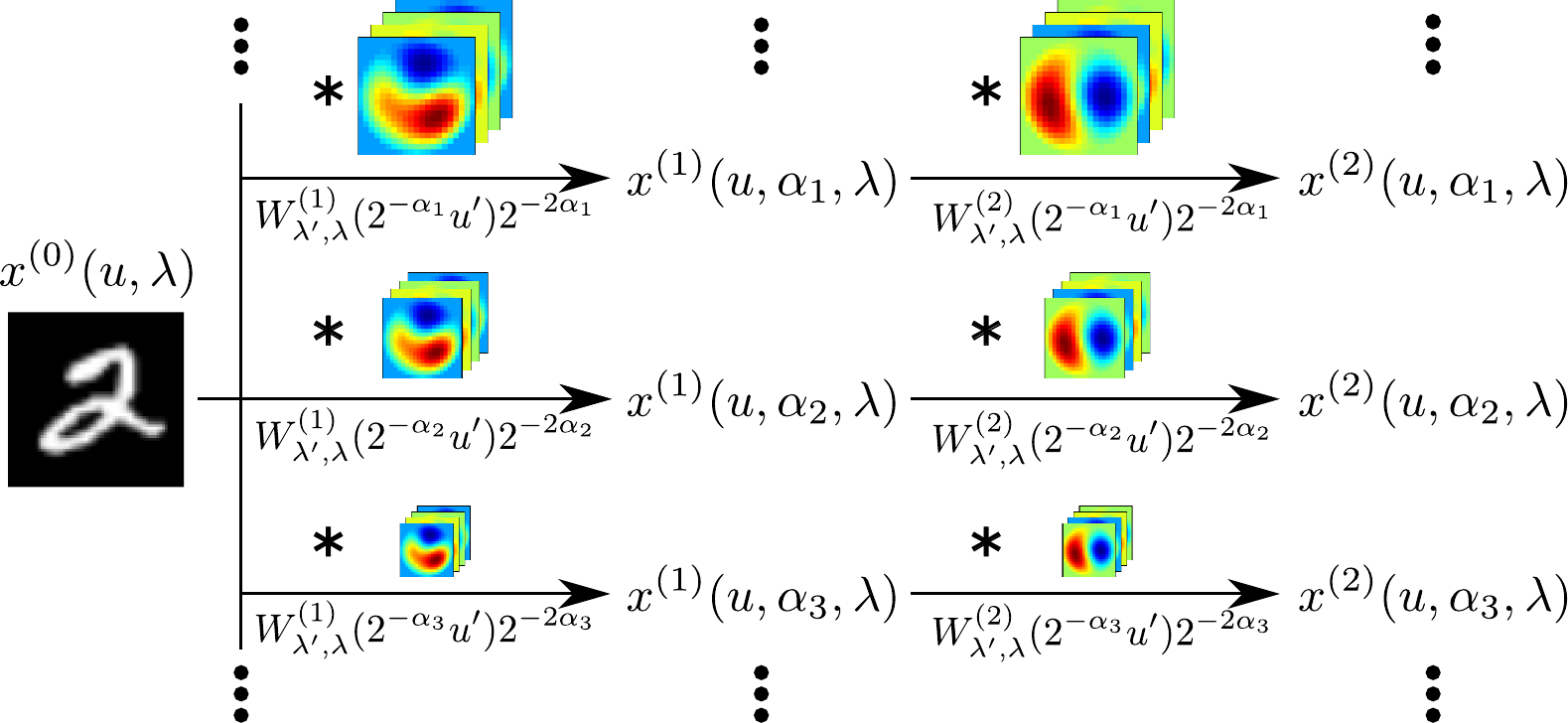}
        \caption{\small A special case of $\cst$-equivariant CNN with (multiscale) spatial convolutions. }
        \label{fig:secnn_special}
    \end{subfigure}
     ~
    \begin{subfigure}[t]{0.49\textwidth}
        \includegraphics[width=\textwidth]{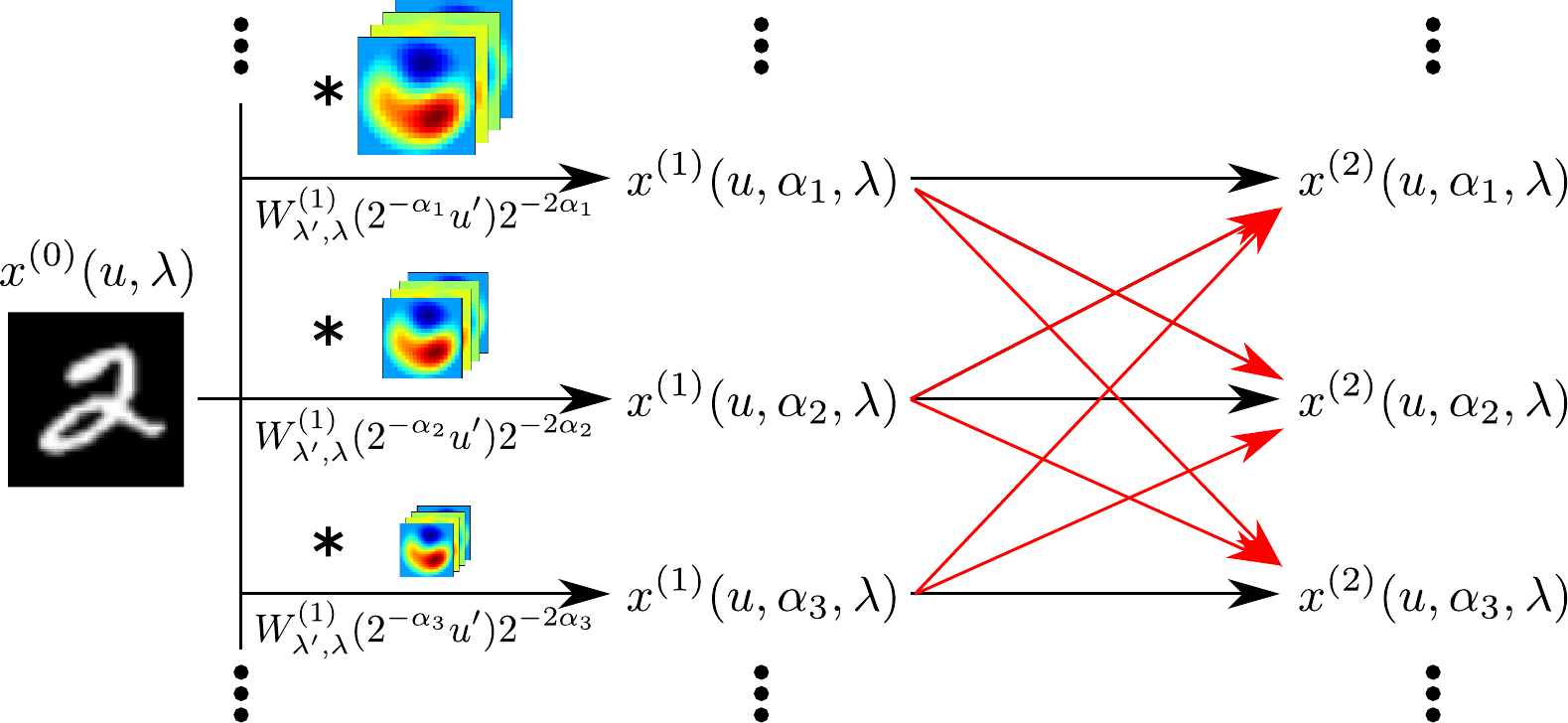}
        \caption{\small The general case of $\cst$-equivariant CNN with joint convolutions (Theorem~\ref{thm:equivariance}). }
        \label{fig:secnn_general}
    \end{subfigure}
    \vspace{-.5em}
  \caption{\small (a) A special case of $\cst$-equivariant CNN with only (multiscale) spatial convolutions. The previous works on $\cst$-equivariant CNNs \cite{Kim_2014_scale, Marcos_2018_scale, Xu_2014_scale, ghosh2019scale} are all variants of this architecture. (b) The general case of $\cst$-equivariant CNN with joint convolutions (Theorem~\ref{thm:equivariance}) where information transfers among different scales. See Remark~\ref{rmk:delta} for more explanation.}
  \label{fig:architecture}
\end{figure}

Inspired by \cite{Cheng_2018_rotdcf}, \cite{Weiler_2018_learning}, and \cite{cohen2018general}, we consider $\cst$-equivariant CNNs with an extra index $\alpha\in\cs$ for the the scaling group $\cs\cong \R$: for each $l\ge 1$, the $l$-th layer output is denoted as  $x^{(l)}(u,\alpha,\lambda)$, where $u\in\R^2$ is the spatial position, $\alpha\in\cs$ is the scale index, and $\lambda\in [M_l] \coloneqq \left\{1, \ldots, M_l\right\}$ corresponds to the unstructured channels.  We use the continuous model  for formal derivation, i.e., the images and feature maps have continuous spatial and scale indices. In practice, the images are discretized on a Cartesian grid, and the scales are computed only on a discretized finite interval (c.f. Section~\ref{sec:implementation}.) We define the group action $T_{\beta,v}$ on the $l$-th layer output as a scaling-translation in space as well as a shift in the scale channel, which corresponds to the regular representation of the group $\cst$ on the space of $l$-th layer feature maps \cite{Cohen_2016_group,cohen2018general}:
\begin{align}
  \label{eq:group-action2}
  T_{\beta,v}x^{(l)}(u,\alpha,\lambda) \coloneqq x^{(l)}\left(2^{-\beta}(u-v),\alpha-\beta,\lambda\right), \quad \forall~l\ge 1.
\end{align}
A feedforward neural network is said to be $\cst$-equivariant (under the group actions $D_{\beta, v}$ and $T_{\beta, v}$) if
\begin{align}
  \label{eq:scale-equivariant}
  x^{(l)}[D_{\beta, v} x^{(0)}]  = T_{\beta, v}x^{(l)}[x^{(0)}], \quad \forall~l\ge 1,
\end{align}
where we slightly abuse the notation $x^{(l)}[x^{(0)}]$ to denote the $l$-th layer output given the input $x^{(0)}$. The following Theorem shows that $\cst$-equivariance \eqref{eq:scale-equivariant} is achieved if and only if joint convolutions are conducted over $\cs\times \R^2$ as in \eqref{eq:1st-layer} and \eqref{eq:lth-layer}.

\begin{thm}
  \label{thm:equivariance}
A feedforward neural network with an extra index $\alpha\in \cs$ for layerwise output is $\cst$-equivariant (under the group actions $D_{\beta, v}$ and $T_{\beta,v}$) if and only if the layerwise operations are defined as \eqref{eq:1st-layer} and \eqref{eq:lth-layer}:
\begin{align}
  \label{eq:1st-layer}
  x^{(1)}[x^{(0)}](u,\alpha, \lambda) &= \sigma\left( \sum_{\lambda'}\int_{\R^2}2^{-2\alpha} x^{(0)}(u+u', \lambda') W_{\lambda', \lambda}^{(1)}\left(2^{-\alpha}u'\right)du'+b^{(1)}(\lambda)\right),\\ 
  \nonumber
  x^{(l)}[x^{(l-1)}](u,\alpha, \lambda) &= \sigma\left( \sum_{\lambda'}\int_{\R^2}\int_{\R} 2^{-2\alpha} x^{(l-1)}(u+u', \alpha+\alpha', \lambda') \right. \cdot \\ 
  \label{eq:lth-layer}
  &\quad \quad \quad \quad \quad \quad \quad \left. W_{\lambda', \lambda}^{(l)}\left(2^{-\alpha}u', \alpha'\right)d\alpha' du'+b^{(l)}(\lambda) \right), \quad \forall l>1,
\end{align}
where $\sigma:\R\to\R$ is a pointwise nonlinear function, $W_{\lambda',\lambda}^{(1)}(u)$ is the spatial convolutional filter in the first layer with output channel $\lambda$ and input channel $\lambda'$, and $W_{\lambda', \lambda}^{(l)}(u, \alpha)$ is the space-scale joint convolutional filter for layer $l>1$.
\end{thm}
We defer the proof of Theorem~\ref{thm:equivariance}, as well as those of other theorems, to the appendix. We note that the joint-convolution in Theorem~\ref{thm:equivariance} is a generalization of the group convolution proposed by \cite{Cohen_2016_group} to a non-compact group $\cst$ in the continuous setting, which is known to be necessary and sufficient to achieve equivariance under regular representations of the group \cite{cohen2018general}.
\begin{rmk}
  \label{rmk:delta}
When the convolutional filter $W_{\lambda',\lambda}^{(l)}(u, \alpha)$ takes the special form $W_{\lambda',\lambda}^{(l)}(u,\alpha) = V_{\lambda',\lambda}^{(l)}(u)\delta(\alpha)$, where $\delta$ is the Dirac delta function, the joint convolution \eqref{eq:lth-layer} over $\R^2\times \cs$ reduces to only a (multiscale) spatial convolution
\begin{align}
  \label{eq:special_case}
  x^{(l)}[x^{(l-1)}](u,\alpha, \lambda) = \sigma\left( \sum_{\lambda'}\int_{\R^2} x^{(l-1)}(u+u', \alpha, \lambda') V_{\lambda', \lambda}^{(l)}\left(2^{-\alpha}u'\right)2^{-2\alpha} du'+b^{(l)}(\lambda)\right),
\end{align}
i.e., the feature maps at different scales do not transfer information among each other (see Figure~\ref{fig:secnn_special}). The previous works \cite{Kim_2014_scale, Marcos_2018_scale, Xu_2014_scale, ghosh2019scale} on $\cst$-equivariant CNNs are all based on this special case of Theorem~\ref{thm:equivariance}.
\end{rmk}

Although the  joint convolutions \eqref{eq:lth-layer} on $\R^2\times \cs$ provide the most general way of imposing $\cst$-equivariance under $D_{\beta,v}$ and $T_{\beta,v}$, they unfortunately also incur a significant increase in the model size and computational burden. 
Following the idea of \cite{Cheng_2018_rotdcf} and \cite{dcfnet}, we address this issue by taking a truncated decomposition of the convolutional filters under a pre-fixed separable basis, which will be discussed in detail in the next section.

\subsection{Separable Basis Decomposition}
\label{sec:separable}

We consider decomposing the convolutional filters $W^{(l)}_{\lambda',\lambda}(u, \alpha)$ under the product of two orthogonal function bases, $\left\{\psi_k(u) \right\}_k$ and  $\left\{\varphi_m(\alpha)\right\}_m$, which are the eigenfunctions of the  Dirichlet Laplacian on, respectively, the rectangle $D = [-1, 1]^2\subset \R^2$ and $I_\alpha=[-1,1]$, i.e., 
\begin{align}
  \label{eq:def-fb}
  \left\{
  \begin{aligned}
    \Delta \psi_k = -\mu_k \psi_k \quad &\text{in}~ D,\\
    \psi_k = 0 \quad &\text{on}~ \partial D,
  \end{aligned}\right.
                       \quad\text{and}\quad\left\{
  \begin{aligned}
    &\varphi_m'' = -\nu_m \varphi_m \quad \text{in}~ I_\alpha=[-1, 1]\\
    &\varphi_m(-1) = \varphi_m(1) = 0.
  \end{aligned}\right.
\end{align}
\begin{rmk}
  The choice of the spatial function basis $\left\{\psi_k(u) \right\}_k$ is not unique. For example, one can also choose $\psi_k(u)$ to be the eigenfunctions of the Dirichlet Laplacian on the unit disk, i.e., the Fourier-Bessel basis \cite{Abramowitz_1965_handbook} considered by \cite{Cheng_2018_rotdcf}. The only requirement for $\left\{\psi_k(u) \right\}_k$ is that it is an orthogonal basis vanishing on the boundary of the domain---a property we will use later in the proof of deformation stability of the equivariant representation (Theorem~\ref{thm:regularity_final}) in Section~\ref{sec:stability}. Empirically, we found $\psi_k(u) = \psi_k(u_1, u_2)$ defined in \eqref{eq:def-fb}, which is separable in the two spatial coordinates, to consistently achieve better results.

  Compared to $\psi_k(u)$, the function basis in scale $\left\{\varphi_m(\alpha) \right\}_m$ is less restrictive, as the choice of which does not contribute to the stability analysis for spatial deformation in Theorem~\ref{thm:regularity_final}. However, as the joint convolutional filters $W^{(l)}_{\lambda',\lambda}(u, \alpha)$ are assumed to be compactly supported in both space and scale (otherwise it is hard to implement in practice), it is natural to choose a compactly supported orthogonal function basis $\{\varphi_m(\alpha)\}_m$ for scale.
\end{rmk}

In the continuous formulation, the spatial ``pooling'' operation is equivalent to rescaling the convolutional filters in space. We thus assume, without loss of generality, that the convolutional filters are compactly supported on a rescaled domain as follows
\begin{align}
  W^{(1)}_{\lambda',\lambda}\in C_c(2^{j_1}D), \quad \text{and}\quad W^{(l)}_{\lambda',\lambda}\in C_c(2^{j_l}D\times I_\alpha), ~~\forall~l>1.
\end{align}
Let  $\psi_{j,k}(u) \coloneqq 2^{-2j}\psi_k(2^{-j}u)$, then  $W_{\lambda',\lambda}^{(l)}$ can be decomposed under $\left\{\psi_{j_l,k} \right\}_k$ and $\left\{\varphi_{m} \right\}_m$ as
\begin{align}
  \label{eq:joint-bases}
  \hspace{-.5em}W_{\lambda',\lambda}^{(1)}(u) = \sum_{k}a_{\lambda',\lambda}^{(1)}(k)\psi_{j_1,k}(u),~ W_{\lambda',\lambda}^{(l)}(u,\alpha) = \sum_m\sum_{k}a_{\lambda',\lambda}^{(l)}(k,m)\psi_{j_l,k}(u)\varphi_m(\alpha), ~l>1
\end{align}
where $a_{\lambda',\lambda}^{(1)}(k)$ and $a_{\lambda',\lambda}^{(l)}(k,m)$ are the expansion coefficients of the filters. During training, the basis functions are fixed, and only the expansion coefficients are updated. In practice, we truncate the expansion to only low-frequency components (i.e.,  $a_{\lambda',\lambda}^{(l)}(k,m)$ are non-zero only for $k\in[K]$, $m\in [K_\alpha]$), which are kept as the trainable parameters. Similar idea has also been considered in the prior works \cite{dcfnet,Cheng_2018_rotdcf,jacobsen2016structured}. Since $\psi_k$ and $\varphi_m$ are the separable eigenfunctions of the Dirichlet Laplacian, standard convergence results for generalized Fourier series apply, e.g., if $W_{\lambda',\lambda}^{(l)}\in C^{p}(D\times I_\alpha)$, then the truncated expansion converges uniformly to $W_{\lambda',\lambda}^{(l)}$ at rate $O(\log(KK_\alpha)/(K^pK_\alpha^p))$. We call the resulting model Scaling-translation-equivariant Network with Decomposed Convolutional Filters (ScDCFNet).

Truncating the filter expansion leads directly to a reduction of network parameters and computational burden. More specifically, let us compare the $l$-th convolutional layer \eqref{eq:lth-layer} of a $\cst$-equivariant CNN with and without truncated basis decomposition:

\textbf{Number of trainable parameters:} Suppose the filters $W^{(l)}_{\lambda',\lambda}(u,\alpha)$ are discretized on a Cartesian grid of size $L\times L\times L_\alpha$. The number of trainable parameters at the $l$-th layer of an $\cst$-equivariant CNN without basis decomposition is $L^2L_\alpha M_{l-1}M_{l}$. On the other hand, in an ScDCFNet with truncated basis expansion up to $K$ leading coefficients for $u$ and $K_\alpha$ coefficients for $\alpha$, the number of parameters is instead $K K_\alpha M_{l-1}M_l$. Hence a reduction to a factor of $K K_\alpha/L^2 L_\alpha$ in trainable parameters is achieved for ScDCFNet via truncated basis decomposition. In particular, if $L = 5, L_\alpha = 5, K = 8$, and $K_\alpha = 3$, then the number of parameters is reduced to $(8\times 3)/(5^2\times 5)= 19.2\%$.

  \begin{rmk}
    We want to point out that even though the number of trainable parameters has been reduced after truncated expansion, the memory usage of the entire network remains the same. The reason is that the bottleneck of memory consumption in a deep network is the storage of the feature maps instead of the trainable weights.
  \end{rmk}

\textbf{Computational cost:} Suppose the size of the input $x^{(l-1)}(u,\alpha,\lambda)$ and output $x^{(l)}(u,\alpha,\lambda)$ at the $l$-th layer are, respectively, $H\times W\times N_s\times M_{l-1}$ and $H\times W\times N_s\times M_l$, where $H\times W$ is the spatial dimension, $N_s$ is the number of scale channels, and $M_{l-1}$($M_{l}$) is the number of the unstructured input (output) channels. Let the filters $W^{(l)}_{\lambda',\lambda}(u,\alpha)$ be discretized on a Cartesian grid of size $L\times L\times L_\alpha$. The following proposition shows that, compared to a regular $\cst$-equivariant CNN, the computational cost in a forward pass of ScDCFNet is reduced again to a factor of $K K_\alpha/L^2 L_\alpha$.
\begin{prop}
  \label{prop:computational-cost}
  Assume $M_l\gg L^2, L_\alpha$, i.e., the number of the output channels is much larger than the size of the convolutional filters in $u$ and $\alpha$,  then the computational cost of an ScDCFNet is reduced to a factor of $K K_\alpha/L^2 L_\alpha$ when compared to a $\cst$-equivariant CNN without basis decomposition.
\end{prop}

\section{Representation Stability of ScDCFNet to Input Deformation}
\label{sec:stability}

Apart from reducing the model size and computational burden, we demonstrate in this section that truncating the filter decomposition has the further benefit of improving  deformation robustness of the equivariant representation, i.e., the equivariance relation \eqref{eq:scale-equivariant} still approximately holds true even if the spatial scaling of the input $D_{\beta,v}x^{(0)}$ is contaminated by a local deformation. The analysis is motivated by the fact that scaling transformations are rarely perfect in practice---they are typically subject to local distortions such as changing view angle or numerical discretization. To quantify the distance between different feature maps at each layer, we define the  norm of $x^{(l)}$ as
\begin{align}
  \label{eq:norms}
  \hspace{-.5em}\|x^{(0)}\|^2 = \frac{1}{M_0}\sum_{\lambda = 1}^{M_0}\int\left|x^{(0)}(u,\lambda) \right|^2du, ~~ \|x^{(l)}\|^2 = \sup_{\alpha\in \R}\frac{1}{M_l}\sum_{\lambda = 1}^{M_l}\int\left|x^{(l)}(u,\alpha, \lambda) \right|^2du, ~ l\ge 1.
\end{align}
\begin{rmk}
  \label{rmk:l_inf}
  The definition of $\|x^{(l)}\|$ is different from that of RotDCFNet \cite{Cheng_2018_rotdcf}, where an $L^2$ norm is taken for the $\alpha$ index as well. The reason why we adopt the $L^\infty$ norm for $\alpha$ in \eqref{eq:norms} is that $x^{(l)}$ is typically not $L^2$ in $\alpha$, since  the scaling group $\cs$, unlike $SO(2)$, has infinite Haar measure.
\end{rmk}

We next quantify the representation stability of ScDCFNet under three mild assumptions on the convolutional layers and input deformations. First,

\textbf{(A1)} The pointwise nonlinear activation $\sigma:\R \to \R$ is non-expansive, i.e., $|\sigma(x)-\sigma(y)|\le |x-y|$. For example, the rectified linear unit (ReLU) satisfies this property.

Next, we need a bound on the convolutional filters under certain norms. For each $l\ge 1$, define $A_l$ as 
\begin{align}
  \label{eq:A_l}
  \left\{
  \begin{aligned}
    &A_1 \coloneqq \pi \max \left\{\sup_{\lambda}\sum_{\lambda'=1}^{M_0}\|a^{(1)}_{\lambda',\lambda}\|_{\mu}, ~\frac{M_0}{M_1}\sup_{\lambda'}\sum_{\lambda=1}^{M_1}\|a^{(1)}_{\lambda',\lambda}\|_{\mu} \right\},\\
    &A_l \coloneqq \pi \max \left\{\sup_{\lambda}\sum_{\lambda'=1}^{M_{l-1}}\sum_{m}\|a^{(l)}_{\lambda',\lambda}(\cdot, m)\|_{\mu}, ~\frac{2M_{l-1}}{M_l}\sum_m\sup_{\lambda'}\sum_{\lambda=1}^{M_l}\|a^{(l)}_{\lambda',\lambda}(\cdot,m)\|_{\mu} \right\},
  \end{aligned}\right.
\end{align}
where the weighted $l^2$-norm $\|a\|_{\mu}$ of a sequence $\left\{a(k)\right\}_{k\ge 0}$ is defined as $\|a\|_{\mu}^2\coloneqq \sum_k \mu_ka(k)^2$, where $\mu_k$ is the $k$-th eigenvalue of the Dirichlet Laplacian on $[-1, 1]^2$ defined in \eqref{eq:def-fb}. We next assume that each $A_l$ is bounded:

\textbf{(A2)} For all $l\ge 1$, $A_l\le 1$.

The boundedness of $A_l$ is facilitated by truncating the basis decomposition to only low-frequency components (small $\mu_k$), which is one of the key idea of ScDCFNet explained in Section~\ref{sec:separable}. After a proper initialization of the trainable coefficients, (A2) can generally be satisfied. The assumption (A2) implies several bounds on the convolutional filters at each layer (c.f. Lemma~\ref{lemma:Al-bounds-everything} in the appendix), which, combined with (A1), guarantees that an ScDCFNet is layerwise non-expansive:
\begin{prop}
\label{prop:nonexpansive}
Under the assumption (A1) and (A2), an ScDCFNet satisfies the following.
\begin{enumerate}[label=(\alph*)]
\item For any $l\ge 1$, the mapping of the $l$-th layer, $x^{(l)}[\cdot]$ defined in \eqref{eq:1st-layer} and \eqref{eq:lth-layer}, is non-expansive, i.e.,
  \begin{align}
    \|x^{(l)}[x_1]-x^{(l)}[x_2]\|\le \|x_1- x_2\|, \quad \forall x_1, x_2.
  \end{align}
\item Let $x_0^{(l)}$ be the $l$-th layer output given a zero bottom-layer input, then $x_0^{(l)}(\lambda)$ depends only on $\lambda$.
\item Let $x_c^{(l)}$ be the centered version of $x^{(l)}$ after removing $x_0^{(l)}$, i.e., $x_c^{(0)}(u,\lambda) \coloneqq x^{(0)}(u,\lambda) - x_0^{(0)}(\lambda) = x^{(0)}(u,\lambda)$, and $x_c^{(l)}(u,\alpha, \lambda)\coloneqq x^{(l)}(u,\alpha,\lambda) - x_0^{(l)}(\lambda), ~\forall l \ge 1$, then $\|x_c^{(l)}\|\le \|x_c^{(l-1)}\|, ~\forall l\ge 1$. As a result, $\|x_c^{(l)}\| \le \|x_c^{(0)}\| = \|x^{(0)}\|$.
\end{enumerate}
\end{prop}

Finally, we make an assumption on the input deformation modulo a global scale change. Given a $C^2$ function $\tau:\R^2\to\R^2$, the spatial deformation $D_\tau$ on the feature maps $x^{(l)}$ is defined as
\begin{align}
  \label{eq:spatial-deformation}
  D_\tau x^{(0)}(u,\lambda) = x^{(0)}(\rho(u), \lambda),\quad\text{and}\quad D_\tau x^{(l)}(u, \alpha,\lambda) = x^{(l)}(\rho(u), \alpha, \lambda), \quad l\ge 1,
\end{align}
where $\rho(u) = u-\tau (u)$. We assume a small local deformation on the input:

\textbf{(A3)} $|\nabla \tau|_{\infty}\coloneqq \sup_u\|\nabla\tau (u)\|<1/5$, where $\|\cdot\|$ is the operator norm.

The following theorem demonstrates the representation stability of an ScDCFNet to input deformation modulo a global scale change.

\begin{thm}
  \label{thm:regularity_final}
  Let $D_{\tau}$ be a small spatial deformation defined in \eqref{eq:spatial-deformation}, and let $D_{\beta, v}, T_{\beta,v}$ be the group actions corresponding to an arbitrary scaling $2^{-\beta}\in \R_+$ centered at $v\in\R^2$ defined in \eqref{eq:group-action1} and \eqref{eq:group-action2}. In an ScDCFNet satisfying (A1), (A2), and (A3), we have, for any $L$,
  \begin{align}
    \label{eq:regularity_final}
    \left\| x^{(L)}[D_{\beta, v}\circ D_\tau x^{(0)}]-T_{\beta, v}x^{(L)}[x^{(0)}] \right\| \le 2^{\beta+1}\left( 4L|\nabla \tau|_\infty + 2^{-j_L}|\tau|_\infty  \right)\|x^{(0)}\|.
  \end{align}
\end{thm}

Theorem~\ref{thm:regularity_final} gauges how approximately equivariant ScDCFNet is if the input undergoes not only a scale change $D_{\beta, v}$ but also a nonlinear spatial deformation $D_\tau$,  which is important both in theory and in practice because the scaling of an object is rarely perfect in reality. However, Theorem~\ref{thm:regularity_final} only considers the stability of the representation under the ideal setting where layerwise feature maps $x^{(l)}(u,\alpha,\lambda)$ are computed (and stored) for all scales $\alpha\in \R$. For practical implementation (to be discussed in more detail in Section~\ref{sec:implementation}), the scale channel $\cs\cong\R$ needs to be truncated to a finite interval $I = [-T, T]\subset \cs$, i.e.,  $x^{(l)}(u,\alpha,\lambda)$ is only computed for $\alpha\in I$, which unavoidably destroys the global scaling symmetry---similar to the fact that truncating an image to a finite spatial support destroys translation symmetry. In practice, given the truncated $l$-th layer feature map $x^{(l-1)}(u, \alpha,\lambda)$ computed only for $\alpha\in I$, one typically first conducts a zero-padding to $x^{(l-1)}(u, \alpha,\lambda)$ in the scale channel before performing the convolution in scale \cite{worrall2019deep,sosnovik2019scale}. This, however, leads to a significant boundary ``leakage'' effect which destroys equivariance of the representation (see Section~\ref{sec:experiments} for detailed empirical examination of such boundary effect.) We thus need to find a way to alleviate such  boundary ``leakage'' issue and analyze its efficacy by studying the equivariance error \eqref{eq:regularity_final} after a scale channel truncation.

  The idea is very simple: after taking a closer look at the definition of the first layer operation \eqref{eq:1st-layer}, one can notice that, ignoring the bias $b^{(1)}(\lambda)$ and the nonlinear operator $\sigma$, it is essentially the convolution of $x^{(0)}$ with an $L^1$-normalized kernel $2^{-2\alpha}W_{\lambda', \lambda}^{(1)}\left(2^{-\alpha}u'\right)$:
  \begin{align}
    x^{(1)}[x^{(0)}](u,\alpha, \lambda) &= \sigma\left( \sum_{\lambda'}\int_{\R^2}2^{-2\alpha} x^{(0)}(u+u', \lambda') W_{\lambda', \lambda}^{(1)}\left(2^{-\alpha}u'\right)du'+b^{(1)}(\lambda)\right)\\ \label{eq:mollifier}
                                        &= \sigma\left( \sum_{\lambda'} x^{(0)}(\cdot, \lambda')* W_{\lambda', \lambda, \alpha}^{(1)}(u)+b^{(1)}(\lambda)\right),
  \end{align}
  where $W_{\lambda', \lambda, \alpha}^{(1)}(u) \coloneqq 2^{-2\alpha}W_{\lambda', \lambda}^{(1)}\left(2^{-\alpha}u'\right)$ forms a (yet to be normalized) mollifier, i.e., an approximation to identity, in $\R^2$ \cite{evans10}. Based on the  basic properties of mollifiers \cite{evans10}, if $x^{(0)}(\cdot, \lambda')$ is a continuous function of $u$ in $\R^2$, then
  \begin{align}
    \lim_{\alpha\to-\infty}x^{(1)}[x^{(0)}](u,\alpha, \lambda) &= \lim_{\alpha\to-\infty}\sigma\left( \sum_{\lambda'} x^{(0)}(\cdot, \lambda')* W_{\lambda', \lambda, \alpha}^{(1)}(u)+b^{(1)}(\lambda)\right)\\ \label{eq:mollifier_convergence}
    & = \sigma\left( \sum_{\lambda'} A_{\lambda',\lambda}x^{(0)}(u, \lambda')+b^{(1)}(\lambda)\right),
  \end{align}
  where $A_{\lambda',\lambda} = \int_{\R^2}W^{(1)}_{\lambda',\lambda}(u)du$. In particular, \eqref{eq:mollifier_convergence} implies that the limit of $x^{(1)}[x^{(0)}](u,\alpha, \lambda)$ exists as $\alpha\to -\infty$. This observation motivates us to consider the one-sided-replicate-padding in the scale channel (i.e., extending the truncated scale channel $I = [-T, T]$ beyond the left end point $\alpha = -T$ according to Neumann boundary condition.) More specifically, the layerwise operations are now defined as follows: for the first layer,
  \begin{align}
    \label{eq:1st-layer-truncated}
    \tilde{x}^{(1)}[x^{(0)}](u,\alpha,\lambda) = \left\{
    \begin{aligned}
      &x^{(1)}[x^{(0)}](u, \alpha, \lambda), \quad && \alpha \in [-T, T],\\
      & x^{(1)}[x^{(0)}](u, -T, \lambda), \quad && \alpha \in (-\infty, -T],
    \end{aligned}\right.
  \end{align}
  i.e., the computation remains the same as \eqref{eq:1st-layer} for $\alpha\in I=[-T, T]$, and the value $x^{(1)}[x^{(0)}](u, \alpha, \lambda)$ at the left end point $\alpha = -T$ is extended beyond the truncated scale interval $I$ for $\alpha\le -T$ such that the next layer scale convolution can be computed on $I$. Similarly, the computations of the subsequent layer are
  \begin{align}
    \label{eq:lth-layer-truncated}
    \tilde{x}^{(l)}[\tilde{x}^{(l-1)}](u,\alpha,\lambda) = \left\{
    \begin{aligned}
      &x^{(l)}[\tilde{x}^{(l-1)}](u, \alpha, \lambda), \quad && \alpha \in [-T, T],\\
      & x^{(l)}[\tilde{x}^{(l-1)}](u, -T, \lambda), \quad && \alpha \in (-\infty, -T].
    \end{aligned}\right.
  \end{align}

  The following theorem quantifies the extra equivariance error incurred from scale channel truncation after the one-sided-replicate-padding \eqref{eq:1st-layer-truncated} and \eqref{eq:lth-layer-truncated} is adopted.

  \begin{thm}
  \label{thm:regularity_final_truncation}
  Under the same assumption of Theorem~\ref{thm:regularity_final}, if the layerwise operations are defined as \eqref{eq:1st-layer-truncated} and \eqref{eq:lth-layer-truncated}, given an input $x^{(0)}(\cdot, \lambda) \in H^1(\R^2)$ (the Sobolev space of functions on $\R^2$ with square-integrable first-order derivatives), we have, for any $L$,
  \begin{align}
    \label{eq:regularity_final_truncation}
    \left\| \tilde{x}^{(L)}[D_{\beta, v}\circ D_\tau x^{(0)}]-T_{\beta, v}\tilde{x}^{(L)}[x^{(0)}] \right\| \le 2^{\beta+1}\left( 4L|\nabla \tau|_\infty + 2^{-j_L}|\tau|_\infty  \right)\|x^{(0)}\| + O(2^{-T}),
  \end{align}
  where the supremum over $\alpha$ in the definition of $\|\cdot\|$ on the left hand side of \eqref{eq:regularity_final_truncation} is taken for $\alpha \le \min(T, T-\beta)$, the common scale domain of the feature maps $\tilde{x}^{(L)}[D_{\beta, v}\circ D_\tau x^{(0)}]$ and $T_{\beta, v}\tilde{x}^{(L)}[x^{(0)}]$, i.e., 
  \begin{align}\nonumber
    & \left\| \tilde{x}^{(L)}[D_{\beta, v}\circ D_\tau x^{(0)}]-T_{\beta, v}\tilde{x}^{(L)}[x^{(0)}] \right\|^2 \\
    = &\sup_{\alpha\le \min(T, T-\beta)}\frac{1}{M_L}\sum_{\lambda = 1}^{M_L}\int\left| \tilde{x}^{(L)}[D_{\beta, v}\circ D_\tau x^{(0)}]-T_{\beta, v}\tilde{x}^{(L)}[x^{(0)}]\right|^2(u,\alpha, \lambda)du.
  \end{align}
  On the contrary, if zero-padding in the scale channel is adopted (which is typically the case such as \cite{worrall2019deep,sosnovik2019scale}) instead of the one-sided-replicate-padding \eqref{eq:1st-layer-truncated} and \eqref{eq:lth-layer-truncated}, the scale channel truncation error in \eqref{eq:regularity_final_truncation} will be $O(1)$ instead of $O(2^{-T})$.
\end{thm}

Theorem~\ref{thm:regularity_final_truncation} demonstrates that the layerwise operations defined in \eqref{eq:1st-layer-truncated} and \eqref{eq:lth-layer-truncated} significantly alleviate the unavoidable boundary ``leakage'' effect incurred from scale channel truncation, contributing to the equivariance error \eqref{eq:regularity_final} an exponentially decaying term in $T$, the length of the truncated scale interval. We will empirically demonstrate this theoretical result in Section~\ref{sec:experiments}.


  \section{Implementation}
  \label{sec:implementation}

  We discuss, in this section, the implementation details of ScDCFNet, including scale channel truncation and feature map discretization, basis and filter generation, discrete scale-space joint convolution, and batch-normalization.

  \subsection{Scale Channel Truncation And Feature Map Discretization}
  As mentioned in Section~\ref{sec:stability}, in practice, the layerwise feature maps $x^{(l)}(u,\alpha,\lambda)$ can only be computed on a truncated scale interval $I = [-T, T]\subset \R$, which is discretized into a uniform grid of size $N_s$. Let $H_l, W_l, M_l$, respectively, be the height, width, and the number of unstructured channels of the $l$-th layer feature map $x^{(l)}$, then the input $x^{(0)}(u, \lambda)$ is stored as an array of shape $[M_0, H_0, W_0]$, and the layerwise output $x^{(l)}(u, \alpha, \lambda)$ has shape $[M_l, N_s, H_l, W_l]$.

  \begin{rmk}
    Due to, for example, interpolation and numerical integration, feature map discretization incurs an extra error to the equivariance of representation. Fortunately, the stability analysis in Theorem~\ref{thm:regularity_final} under the continuous setting suggests that some of these errors can be mitigated. For instance, interpolating a rescaled digital image can be modeled as a perfect spatial rescaling followed by a small local distortion, the error induced by which in representation equivariance can be controlled in ScDCFNet thanks to Theorem~\ref{thm:regularity_final}.
  \end{rmk}

  \subsection{Basis And Filter Generation}

  Let $K$ and $K_\alpha$, respectively, be the numbers of the low-frequency components to be kept in the separable basis expansion in the spatial and scale domains. The spatial basis functions together with their rescaled versions $\{2^{-2\alpha}\psi_k(2^{-\alpha}u') \}_{k, \alpha, u'}$ are sampled on a uniform spatial grid of size $L\times L$ and stored as a tensor of shape $[K, N_s, L, L]$; the basis functions in scale $\{\varphi_m(\alpha') \}_{ m, \alpha'}$ supported on the interval $I_\alpha \ni 0$ is sampled on a uniform grid of size $L_\alpha$ and stored as a tensor of shape $[K_\alpha, L_\alpha]$.
  
  For the first layer, the truncated expansion coefficients $\{a^{(1)}_{\lambda', \lambda}(k)\}_{\lambda',\lambda,k}$ forming an array of shape $[M_{0}, M_1, K]$ are the trainable parameters of ScDCFNet at this layer, and the multiscale convolutional filters $\{2^{-2\alpha}W_{\lambda',\lambda}^{(1)}(2^{-\alpha}u')\}_{\lambda',\lambda,\alpha,u'}$ are the linear combinations of the spatial basis functions  $\{2^{-2\alpha}\psi_k(2^{-\alpha}u')\}_{k,\alpha,u'}$ under the coefficients $\{a^{(1)}_{\lambda', \lambda}(k)\}_{\lambda',\lambda,k}$, stored as an array of shape $[M_{0}, M_1, N_s, L, L]$.

  When $l>1$,  the tensor consisting of the $l$-th layer trainable coefficients $\{a^{(l)}_{\lambda', \lambda}(k,m)\}_{\lambda',\lambda,k,m}$ has shape $[M_{l-1}, M_l, K, K_\alpha]$. The joint convolutional filters $\{2^{-2\alpha}W_{\lambda',\lambda}^{(l)}(2^{-\alpha}u', \alpha')\}_{\lambda',\lambda, \alpha, \alpha', u'}$ of this layer are the linear combinations of the separable function bases  $\{2^{-2\alpha}\psi_k(2^{-\alpha}u') \}_{k,\alpha,u'}$ and $\{\varphi_m(\alpha') \}_{m,\alpha'}$ under the coefficients $\{a^{(l)}_{\lambda', \lambda}(k,m)\}_{\lambda',\lambda,k,m}$, i.e.,  they are stored as an array of size $[M_{l-1}, M_l, N_s, L_\alpha, L, L]$ which is the tensor product of the first two arrays followed by a contraction with the third.
  
  \begin{rmk}
    \label{rmk:small_taps}
    Another effective way to reduce the boundary ``leakage'' phenomenon (c.f. Section~\ref{sec:stability}) is to choose a much smaller support $I_\alpha$ for the convolutional filters $W_{\lambda',\lambda}^{(l)}(u', \alpha')$ in scale compared to the truncated scale interval $I$, i.e., $|I_{\alpha}|\ll |I|$, or equivalently, $L_\alpha \ll N_s$, (a similar idea has been explored by \cite{worrall2019deep} and \cite{sosnovik2019scale}.) The efficacy of this will be empirically examined in Section~\ref{sec:experiments}.  
  \end{rmk}

  \subsection{Discrete Scale-Space Joint Convolution}
  Given an input signal $x^{(0)}(u, \lambda)$ of shape $[M_0, H_0, W_0]$, a multiscale discrete spatial convolution, i.e., the discrete counterpart of \eqref{eq:1st-layer} where integrals are replaced by summations, with the filter $\{2^{-2\alpha}W_{\lambda',\lambda}^{(1)}(2^{-\alpha}u')\}_{\lambda',\lambda,\alpha,u'}$ of shape $[M_0, M_1, N_s, L, L]$ is conducted to obtain the first layer feature $x^{(1)}(u,\alpha, \lambda)$ of shape $[M_1, N_s,  H_1, W_1]$. More specifically, the standard 2D convolution is performed on the input with the filter reshaped to $[M_0, M_1N_s, L, L]$, producing an output of shape $[M_1N_s, H_1, W_1]$, which is subsequently resized to $[M_1, N_s,  H_1, W_1]$.

  Starting from the second layer, let  $x^{(l-1)}(u,\alpha,\lambda)$ be the feature of shape $[M_{l-1}, N_s, H_{l-1}, W_{l-1}]$, and $F = \{2^{-2\alpha}W_{\lambda',\lambda}^{(l)}(2^{-\alpha}u', \alpha')\}_{\lambda',\lambda, \alpha, \alpha', u'}$  be the filter of size $[M_{l-1}, M_l, N_s, L_\alpha, L, L]$. The signal $x^{(l-1)}$ is first shifted in the scale channel by $l_\alpha\in [0, L_\alpha-1]$ according to one-sided-replicate-padding (c.f. Section~\ref{sec:stability}), and then convolved with the filter $F[:,:,:,l_\alpha,:,:]$ (after reshaping) to obtain an output tensor of size $[M_l, N_s, H_l, W_l]$. We iterate over all $l_\alpha\in [0, L_\alpha-1]$, and the $l$-th layer feature $x^{(l)}(u,\alpha,\lambda)$ is the sum of the $L_\alpha$ tensors.

  For computer vision tasks where scale-invariant features are preferred, e.g., image classification, a max-pooling in the scale channel is conducted on the last layer feature $x^{(L)}(u,\alpha,\lambda)$ of size $[M_L, N_s,  H_L, W_L]$,  producing a scale-invariant output of shape $[M_L,  H_L, W_L]$. Without explicitly mentioning, such max-pooling in scale is only performed in the last layer.

  \subsection{Batch-Normalization}
  Batch-normalization \cite{Ioffe_2015_batchnorm} accelerates network training by reducing layerwise covariate shift, and it has become an integral part in various CNN architectures. With an extra scale index $\alpha$ in the feature map $x^{(l)}(u,\alpha,\lambda)$ of an ScDCFNet, we need to include $\alpha$ in the normalization in order not to destroy $\cst$-equivariance, i.e., a batch of features $\{x^{(l)}_n(u,\alpha,\lambda)\}_{n=1}^N$ should be normalized as if it were a collection of 3D data (two dimensions for $u$, and one dimension for $\alpha$.)

\section{Numerical Experiments}
\label{sec:experiments}

In this section, we conduct several numerical experiments for the following three purposes.
\begin{enumerate}
\item To demonstrate that ScDCFNet robustly achieves $\cst$-equivariance \eqref{eq:scale-equivariant} in the practical setting where signals are discrete and scale channels are truncated, verifying the theoretical results Theorem~\ref{thm:regularity_final} and Theorem~\ref{thm:regularity_final_truncation}.
\item To illustrate that ScDCFNet significantly outperforms regular CNNs as well as other competing scale-invariant/equivariant networks at a much reduced model size in multiscale image classification.
\item To show that a trained ScDCFNet auto-encoder is able to reconstruct rescaled versions of the input by simply applying group actions on the image codes, demonstrating that ScDCFNet indeed explicitly encodes the input scale information into the representation.
\end{enumerate}

\subsection{Data Sets And Models}
\label{sec:data}

The experiments are conducted on the Scaled MNIST (SMNIST), Scaled Fashion-MNIST (SFashion), and the STL-10 data sets \cite{stl}.

SMNIST and SFashion are built by rescaling the original MNIST \cite{lecun1998gradient} and Fashion-MNIST \cite{Xiao_2017_fashion} images by a factor randomly sampled from a uniform distribution on $[0.3, 1]$.  A zero-padding back to a size of $28 \times 28$ is conducted after the rescaling. If mentioned explicitly, for some experiments, the images are resized to $56\times 56$ for better visualization.

The STL-10 data set is comprised of 5,000 training and 8,000 testing labeled RGB images belonging to 10 classes such as airplane, bird, and car.  The images have a spatial resolution of $96\times 96$, and we do not rescale the images due to the rich scaling variance already existing in the data set.

We compare the performance of our ScDCFNet with other network models that deal with scaling variation within the input data, including the local scale-invariant models LSI-CNN \cite{Kim_2014_scale}, SS-CNN \cite{ghosh2019scale}, and scale-equivariant models SI-CNN \cite{Xu_2014_scale}, SEVF \cite{Marcos_2018_scale}, DSS \cite{worrall2019deep}, and SESN \cite{sosnovik2019scale}.

\subsection{Verification of  $\cst$-Equivariance}
\label{sec:ex1}
\begin{figure}[t]
  \centering
  \includegraphics[width=1\textwidth]{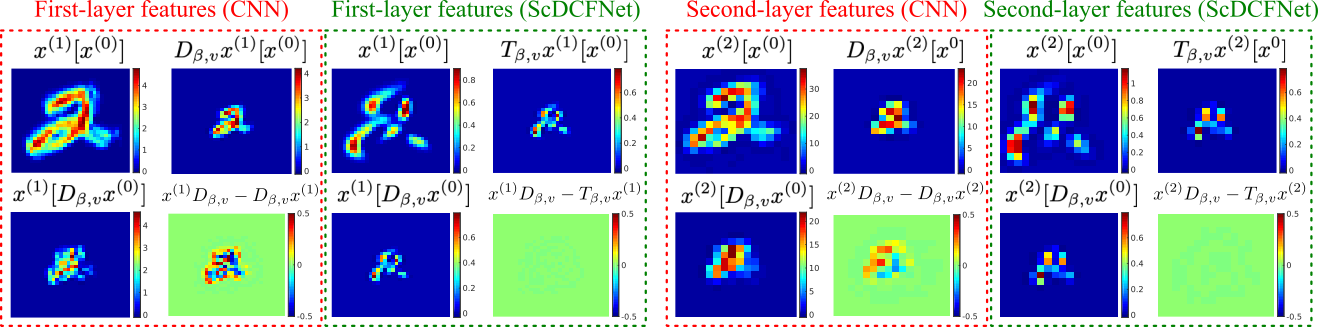}
  \vspace{-.5em}
  \caption{\small Verification of $\cst$-equivariance in Section~\ref{sec:ex1}. Given the original input $x^{(0)}$ and its rescaled version $D_{\beta, v}x^{(0)}$, the four figures in each dashed rectangle are: $x^{(l)}[x^{(0)}]$ ($l$-th layer feature of the original input), $x^{(l)}[D_{\beta,v}x^{(0)}]$ ($l$-th layer feature of the rescaled input), $T_{\beta,v}x^{(l)}[x^{(0)}]$ (rescaled $l$-th layer feature of the original input), and the difference $(x^{(l)}[D_{\beta,v}x^{(0)}]-T_{\beta,v}x^{(l)}[x^{(0)}])$ displayed in a (signal intensity) scale relative to the maximum value of $x^{(l)}[D_{\beta, v}x^{(0)}]$. It is clear that even after numerical discretization, $\cst$-equivariance still approximately holds for ScDCFNet, i.e., $x^{(l)}[D_{\beta,v}x^{(0)}]-T_{\beta,v}x^{(l)}[x^{(0)}]\approx 0$, but not for a regular CNN.}\label{fig:equivariance}
\end{figure}

We first verify that ScDCFNet indeed achieves $\cst$-equivariance \eqref{eq:scale-equivariant}. Specifically, we compare the feature maps of a  two-layer ScDCFNet with randomly generated truncated filter expansion coefficients and those of a regular CNN. The exact architectures are detailed in Appendix~\ref{sec:app_ex1}. Figure~\ref{fig:equivariance} displays the first- and second-layer feature maps of an original image $x^{(0)}$ and its rescaled version $D_{\beta, v}x^{(0)}$ using the two comparing architectures. Feature maps at different layers are rescaled to the same spatial dimension for visualization. The four images enclosed in each of the dashed rectangle correspond to:  $x^{(l)}[x^{(0)}]$ ($l$-th layer feature of the original input), $x^{(l)}[D_{\beta,v}x^{(0)}]$ ($l$-th layer feature of the rescaled input), $T_{\beta,v}x^{(l)}[x^{(0)}]$ (rescaled $l$-th layer feature of the original input, where $T_{\beta,v}$ is understood as $D_{\beta, v}$ for a regular CNN due to the lack of a scale index $\alpha$), and the difference $x^{(l)}[D_{\beta,v}x^{(0)}]-T_{\beta,v}x^{(l)}[x^{(0)}]$. It is clear that even with numerical discretization, which can be modeled as a form of input deformation, ScDCFNet is still approximately $\cst$-equivariant, i.e., $x^{(l)}[D_{\beta,v}x^{(0)}] \approx T_{\beta,v}x^{(l)}[x^{(0)}]$, whereas a regular CNN does not have such a property.

\begin{figure}
  \centering
      \begin{subfigure}[t]{0.4\textwidth}
        \includegraphics[width=\textwidth]{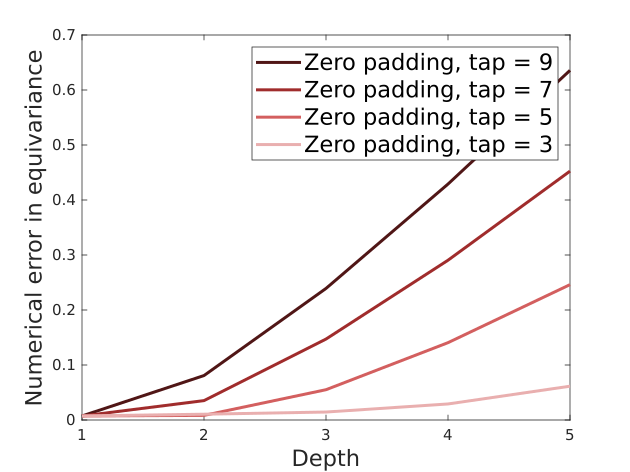}
        \caption{\small Zero-padding for the scale channel. }
        \label{fig:error_depth_zero}
    \end{subfigure}
     ~~~~~
    \begin{subfigure}[t]{0.4\textwidth}
        \includegraphics[width=\textwidth]{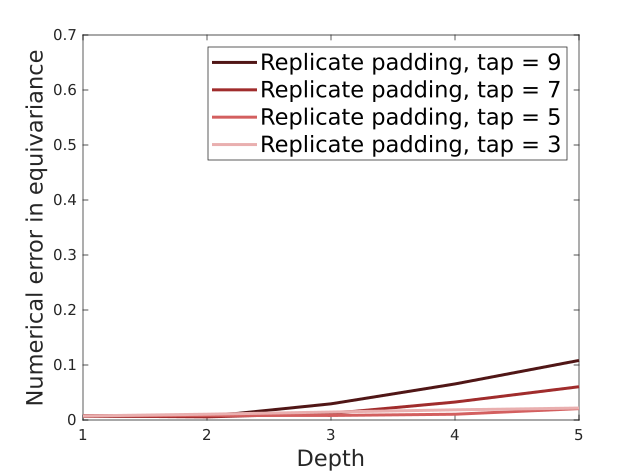}
        \caption{\small One-sided-replicate-padding for the scale channel. }
        \label{fig:error_depth_replicate}
    \end{subfigure}
    \vspace{-.5em}
  \caption{\small The numerical error in equivariance incurred by the boundary ``leakage'' effect after scale channel truncation as a function of network depth. Either (a) zero-padding or (b) one-sided-replicate-padding is used for the convolution in scale. The error is unavoidable as depth becomes larger, but it can be mitigated by (1) using joint convolutional filters with a smaller support $I_\alpha$ in scale (i.e., a smaller number of ``taps'' $L_\alpha$ after discretization), and (2) using a one-sided-replicate-padding instead of zero-padding.}
  \label{fig:error_depth}
\end{figure}

We also examine how the numerical error in equivariance incurred by the the boundary ``leakage'' effect after the scale channel truncation evolves as the network gets deeper. The error in equivariance is measured in a relative $L^2$ sense at a particular scale $\alpha$, i.e.,
\begin{align}
\text{Error} = \|x^{(l)}[D_{\beta,v}x^{(0)}](\cdot, \alpha)-T_{\beta,v}x^{(l)}[x^{(0)}](\cdot, \alpha)\|_{L^2} / \|T_{\beta,v}x^{(l)}[x^{(0)}](\cdot, \alpha)\|_{L^2}.
\end{align}
It is clear from Figure~\ref{fig:error_depth} that the boundary ``leakage'' effect is unavoidable as the network becomes deeper. However, the error can be significantly alleviated by either choosing joint convolutional filters with a smaller support $I_\alpha$ in scale (i.e., the filter size $L_\alpha$ in scale, or the number of ``taps'', is much smaller compared to the number of scale channels $N_s$ in the feature map, (c.f. Remark~\ref{rmk:small_taps})), or using a one-sided-replicate-padding in the  scale channel for the joint convolution instead of the default zero-padding typically adopted in other works such as \cite{worrall2019deep,sosnovik2019scale} (c.f. Theorem~\ref{thm:regularity_final_truncation}.)

\subsection{Multiscale Image Classification}
\label{sec:classification}

We next demonstrate the improved performance of ScDCFNet in multiscale image classification compared to regular CNNs and other scale-invariant or scale-equivariant models.

\subsubsection{SMNIST And SFashion}

  \begin{table}[t]
  \centering
  \scriptsize
  \begin{tabular}{lccccccc}
    \toprule
    &\multicolumn{3}{c}{SMNIST $(28\times 28)$ test accuracy (\%)}&\multicolumn{3}{c}{SMNIST $(28 \times 28)+$ test accuracy (\%)}\\
    \cmidrule(r){2-4} \cmidrule(r){5-7}
    Models   &$N_{\text{tr}} = 2,000$ & $N_{\text{tr}} = 5,000$ &  $N_{\text{tr}} = 10,000$ &$N_{\text{tr}} = 2,000$ & $N_{\text{tr}} = 5,000$ & $N_{\text{tr}} = 10,000$\\
    \midrule
    CNN & $95.01\pm 0.27$ & $96.39\pm 0.12$ & $97.41\pm 0.13$ & $96.36\pm 0.43$ & $97.27\pm 0.14$ & $98.05\pm 0.07$\\
    SS-CNN & $95.43\pm 0.21$ & $96.70\pm 0.20$ & $97.64\pm 0.08$& $96.13\pm 0.25$ & $96.84\pm 0.17$ & $97.98\pm 0.03$\\
    SEVF & $95.09\pm 0.15$ & $96.29\pm 0.15$ & $97.28\pm 0.16$ & $96.06\pm 0.25$ & $96.68\pm 0.09$ & $97.74\pm 0.14$\\
    LSI-CNN & $95.46\pm 0.22$ & $96.64\pm 0.08$ & $97.56\pm 0.13$ & $96.43\pm 0.23$ & $97.17\pm 0.08$ & $97.97\pm 0.05$\\
    SI-CNN & $95.17\pm 0.21$ & $96.58\pm 0.22$ & $97.53\pm 0.12$ & $96.44\pm 0.09$ & $97.21\pm 0.27$ & $98.08\pm 0.09$\\
    DSS & $95.06\pm 0.24$ & $96.42\pm 0.17$ & $97.34\pm 0.13$ & $96.56\pm 0.13$ & $97.24\pm 0.10$ & $98.03\pm 0.06$\\
    SESN & $95.75\pm 0.21$ & $96.87\pm 0.12$ & $97.81\pm 0.13$ & $96.61\pm 0.14$ & $97.30\pm 0.15$ & $98.11\pm 0.09$\\
    \midrule
    ScDCFNet & $\bm{95.87\pm 0.18}$ & $\bm{97.09\pm 0.05}$ & $\bm{97.91\pm 0.08}$ & $\bm{96.68\pm 0.13}$ & $\bm{97.40\pm 0.18}$ & $\bm{98.19\pm 0.07}$\\
    \bottomrule\toprule
    &\multicolumn{3}{c}{SMNIST $(56\times 56)$ test accuracy (\%)}&\multicolumn{3}{c}{SMNIST $(56 \times 56)+$ test accuracy (\%)}\\
    \cmidrule(r){2-4} \cmidrule(r){5-7}
    Models   &$N_{\text{tr}} = 2,000$ & $N_{\text{tr}} = 5,000$ &  $N_{\text{tr}} = 10,000$ &$N_{\text{tr}} = 2,000$ & $N_{\text{tr}} = 5,000$ & $N_{\text{tr}} = 10,000$\\
    \midrule
    CNN & $96.15\pm 0.19$ & $97.20 \pm 0.10$ & $98.01\pm 0.08$ & $97.17\pm 0.12$ & $97.75\pm 0.12$ & $98.32\pm 0.04$\\
    SS-CNN & $96.32\pm 0.12$ & $97.37\pm 0.22$ & $98.08\pm 0.12$& $96.90\pm 0.18$ & $97.57\pm 0.09$ & $98.28\pm 0.14$\\
    SEVF & $96.23\pm 0.17$ & $97.19\pm 0.13$ & $98.01\pm 0.12$ & $96.86\pm 0.14$ & $97.51\pm 0.08$ & $98.20\pm 0.20$\\
    LSI-CNN & $96.43\pm 0.18$ & $97.34\pm 0.17$ & $98.13\pm 0.13$ & $97.11\pm 0.16$ & $97.66\pm 0.07$ & $98.29\pm 0.16$\\
    SI-CNN & $95.97\pm 0.18$ & $97.24\pm 0.12$ & $97.98\pm 0.14$ & $97.16\pm 0.12$ & $97.78\pm 0.06$ & $98.44\pm 0.08$\\
    DSS & $96.10\pm 0.18$ & $97.26\pm 0.10$ & $97.97\pm 0.10$ & $97.19\pm 0.08$ & $97.83\pm 0.10$ & $98.41\pm 0.11$\\
    SESN & $96.50\pm 0.19$ & $97.52\pm 0.10$ & $98.19\pm 0.12$ & $97.21\pm 0.20$ & $97.80\pm 0.11$ & $98.40\pm 0.09$\\
    \midrule
    ScDCFNet & $\bm{96.75\pm 0.18}$ & $\bm{97.61\pm 0.11}$ & $\bm{98.27\pm 0.09}$ & $\bm{97.32\pm 0.08}$ & $\bm{97.85\pm 0.15}$ & $\bm{98.46\pm 0.09}$\\    
    \bottomrule
  \end{tabular}
  \vspace{-.5em}
  \caption{\small Classification accuracy on the SMNIST data set. Models are trained on $N_{\text{tr}}=$ 2K, 5K, or 10K images with spatial resolution $28\times 28$ or $56\times 56$. A plus sign ``+'' is used to denote the presence of scaling data augmentation during training. Test accuracies are reported as mean~$\pm$~std over five independent realizations of the rescaled data set.}  \label{tab:acc_smnist}
\end{table}

\begin{table}[t]
  \centering
  \scriptsize
  \begin{tabular}{lccccccc}
    \toprule
    &\multicolumn{3}{c}{SFashion $(28\times 28)$ test accuracy (\%)}&\multicolumn{3}{c}{SFashion $(28 \times 28)+$ test accuracy (\%)}\\
    \cmidrule(r){2-4} \cmidrule(r){5-7}
    Models   &$N_{\text{tr}} = 2,000$ & $N_{\text{tr}} = 5,000$ &  $N_{\text{tr}} = 10,000$ &$N_{\text{tr}} = 2,000$ & $N_{\text{tr}} = 5,000$ & $N_{\text{tr}} = 10,000$\\
    \midrule
    CNN & $79.88\pm 0.60$ & $82.71\pm 0.19$ & $84.60\pm 0.30$ & $81.82\pm 0.49$ & $83.85\pm 0.51$ & $86.53\pm 0.14$\\
    SS-CNN & $79.24\pm 0.18$ &$82.83\pm 0.49$  & $85.39\pm 0.32$& $80.15\pm 0.31$ & $83.09\pm 0.14$ & $85.89\pm 0.19$\\
    SEVF & $79.01\pm 0.45$ & $81.76\pm 0.40$ & $84.73\pm 0.11$ & $79.47\pm 0.44$ & $82.36\pm 0.30$ & $85.23\pm 0.16$\\
    LSI-CNN & $79.20\pm 0.78$ & $82.58\pm 0.52$  & $85.16\pm 0.14$ & $80.15\pm 0.71$ & $83.10\pm 0.26$ & $86.06\pm 0.16$\\
    SI-CNN & $79.95\pm 0.53$ & $83.36\pm 0.29$ & $85.32\pm 0.22$ & $82.27\pm 0.46$ & $84.09\pm 0.40$ & $86.85\pm 0.15$\\
    DSS & $79.82\pm 0.44$ & $82.90\pm 0.22$ & $84.50\pm 0.51$ & $82.20\pm 0.51$ & $84.04\pm 0.24$ & $86.51\pm 0.27$\\
    SESN & $80.88\pm 0.51$ & $83.78\pm 0.27$ & $85.93\pm 0.28$ & $82.34\pm 0.52$ & $84.26\pm 0.23$ & $86.90\pm 0.27$\\
    \midrule
    ScDCFNet & $\bm{81.32 \pm 0.41}$ & $\bm{84.24\pm 0.35}$ & $\bm{86.19\pm 0.15}$ & $\bm{82.42\pm 0.38}$ & $\bm{84.31\pm 0.30}$ & $\bm{87.10\pm 0.34}$\\
    \bottomrule\toprule
    &\multicolumn{3}{c}{SFashion $(56\times 56)$ test accuracy (\%)}&\multicolumn{3}{c}{SFashion $(56 \times 56)+$ test accuracy (\%)}\\
    \cmidrule(r){2-4} \cmidrule(r){5-7}
    Models   &$N_{\text{tr}} = 2,000$ & $N_{\text{tr}} = 5,000$ &  $N_{\text{tr}} = 10,000$ &$N_{\text{tr}} = 2,000$ & $N_{\text{tr}} = 5,000$ & $N_{\text{tr}} = 10,000$\\
    \midrule
    CNN & $81.17\pm 0.39$ & $84.05\pm 0.12$ & $85.84\pm 0.47$ & $83.35\pm 0.16$ & $85.16\pm 0.22$ & $87.54\pm 0.18$\\
    SS-CNN & $80.97\pm 0.30$ & $83.95\pm 0.33$ & $86.42\pm 0.23$& $81.48\pm 0.75$ & $84.24\pm 0.38$ & $86.69\pm 0.23$\\
    SEVF & $81.22\pm 0.35$ & $84.13\pm 0.27$ & $86.37\pm 0.30$ & $81.88\pm 0.47$ & $84.28\pm 0.32$ & $86.96\pm 0.17$\\
    LSI-CNN & $81.78\pm 0.47$ & $84.49\pm 0.30$ & $86.94\pm 0.34$ & $82.44\pm 0.53$ & $84.71\pm 0.26$ & $87.32\pm 0.16$\\
    SI-CNN & $81.11\pm 0.48$ & $84.18\pm 0.34$ & $86.22\pm 0.17$ & $83.55\pm 0.40$ & $85.18\pm 0.49$ & $87.97\pm 0.16$\\
    DSS & $81.41\pm 0.32$ & $84.15\pm 0.25$ & $85.91\pm 0.23$ & $83.54\pm 0.40$ & $85.44\pm 0.40$ & $87.63\pm 0.07$\\
    SESN & $82.34\pm 0.51$ & $85.03\pm 0.14$ & $86.95\pm 0.23$ & $83.61\pm 0.40$ & $85.40\pm 0.23$ & $87.91\pm 0.13$\\
    \midrule
    ScDCFNet & $\bm{83.01\pm 0.31}$ & $\bm{85.79\pm 0.26}$ & $\bm{87.63\pm 0.15}$ & $\bm{84.00\pm 0.41}$ & $\bm{86.14\pm 0.25}$ & $\bm{88.18\pm 0.17}$\\        
    \bottomrule
  \end{tabular}
  \vspace{-.5em}
  \caption{\small Classification accuracy on the SFashion data set. Models are trained on $N_{\text{tr}}=$ 2K, 5K, or 10K images with spatial resolution $28\times 28$ or $56\times 56$. A plus sign ``+'' is used to denote the presence of scaling data augmentation during training. Test accuracies are reported as mean~$\pm$~std over five independent realizations of the rescaled data set.}  \label{tab:acc_sfashion}
\end{table}

We first test  the comparing models on the SMNIST and SFashion data sets. Five independent realizations of the rescaled data sets are generated according to Section~\ref{sec:data} and kept the same for all networks. Each rescaled data set is split into 10,000 images for training, 2,000 images for evaluation, and 50,000 images for testing. In order to demonstrate the performance of various models in the small data regime, we also report the results trained with 2,000 and 5,000 images.

Following the experimental setup by \cite{ghosh2019scale}, a baseline CNN consisting of three convolutional and two fully-connected layers with batch-normalization is used as a benchmark. The number of output channels of the three convolutional and the first fully-connected layers are set to $[32, 63, 95, 256]$. Convolutional filters of size $7\times 7$ are used in each layer. All comparing networks are built on the same CNN baseline, and the number of trainable parameters are kept almost the same across different models by varying the number of unstructured channels. For equivariant models, a max-pooling in the scale channel is performed only after the final convolutional layer to produce scale-invariant features for classification. For $\cst$-equivariant models achieved by space-scale joint convolution such as SESN and ScDCFNet, the filter size in scale $L_\alpha$ is set to 3. Finally, for ScDCFNet, we set $K = 15$, $K_\alpha = 3$, and $N_s = 5$.

All networks are trained with the Adam optimizer \cite{kingma2014adam} for 60 epochs with a batch size of 128. The initial learning rate is set to 0.01 and scheduled to decrease tenfold after 20 and 40 epochs. We conduct the experiments in 12 different settings, where
\begin{itemize}[noitemsep,topsep=0pt]
\item the input images size is either $28\times 28$ or $56\times 56$;
\item the models are trained with or without scaling data augmentation;
\item the number of training samples $N_{\text{tr}}$ is 2,000, 5,000, or 10,000.
\end{itemize}

We conduct the experiments on five independent realizations of the rescaled data, and report the mean~$\pm$~std of the test accuracy in Table~\ref{tab:acc_smnist} and Table~\ref{tab:acc_sfashion}, where, for instance, $(28\times 28)$ (or $(28\times 28)+$) denotes models are trained on images of size $28\times 28$ without (or with) data augmentation. It can be observed from Table~\ref{tab:acc_smnist} and Table~\ref{tab:acc_sfashion} that ScDCFNet consistently outperforms all comparing methods in every experimental setting, and the improvement in accuracy is especially pronounced in the small data regime without data augmentation.

We want to note that the results of SESN displayed in Table~\ref{tab:acc_smnist} is slightly worse than those reported by \cite{sosnovik2019scale}---this is because even though \cite{sosnovik2019scale} also propose space-scale joint convolution to achieve $\cst$-equivariance, it is however not implemented in their work for the SMNIST experiment, i.e., they set the support of the convolutional filters in scale $L_\alpha$ to 1 instead of 3 (or no ``interscale interaction'' according to the  terminology by \cite{sosnovik2019scale}.) The deterioration in test performance of SESN when $L_\alpha>1$ is in line with the finding by \cite{sosnovik2019scale} on the STL-10 data set that ``interscale interaction'' significantly reduces the accuracy of SESN due to the high equivariance error introduced by the boundary ``leakage'' effect. This is however successfully mitigated in ScDCFNet, verifying again our theoretical result Theorem~\ref{thm:regularity_final_truncation}.

\begin{table}[t]
  \centering
  \scriptsize
  \begin{tabular}{ccccccc}
    \toprule
    \multicolumn{3}{c}{ScDCFNet}   &\multicolumn{2}{c}{SMNIST $(28\times 28)$ accuracy (\%)}&\multicolumn{2}{c}{SMNIST $(28 \times 28)+$ accuracy (\%)}\\
    \cmidrule(r){1-3} \cmidrule(r){4-5} \cmidrule(r){6-7}
    $K_\alpha$ & $K$ & \# Params  &$N_{\text{tr}} = 2,000$ & $N_{\text{tr}} = 5,000$ &  $N_{\text{tr}} = 2,000$ &$N_{\text{tr}} = 5,000$\\
    \midrule
    3 & 15  & $1.00$ & $95.87\pm 0.18$ & $\bm{97.09\pm 0.05}$ & $\bm{96.68\pm 0.13}$ & $\bm{97.40\pm 0.18}$\\
    3 & 12  & $0.80$ & $\bm{95.88\pm 0.21}$ & $97.06\pm 0.07$ & $96.67\pm 0.19$ & $97.38\pm 0.16$\\
    3& 10 & $0.67$ & $95.84\pm 0.18$ & $97.00\pm 0.08$ & $96.49\pm 0.09$ & $97.23\pm 0.12$\\
    3 & 6& $0.40$ & $95.61\pm 0.18$ & $96.89\pm 0.19$ & $96.50\pm 0.15$ & $97.17\pm 0.15$\\
    2& 15 & $0.67$ & $95.77\pm 0.15$ & $97.04\pm 0.05$ & $\bm{96.68\pm 0.31}$ & $97.34\pm 0.08$\\
    2 & 10 & $0.44$ & $95.72\pm 0.23$ & $96.97\pm 0.07$ & $96.52\pm 0.11$ & $97.30\pm 0.13$\\
    2 & 6& $0.27$ & $95.68\pm 0.17$ & $96.92\pm 0.13$ & $96.50\pm 0.14$ & $97.09\pm 0.11$\\
    1 & 15 & $0.33$ & $95.84\pm 0.11$ & $96.94\pm 0.12$ & $96.65\pm 0.19$ & $97.38\pm 0.07$\\
    1 & 10& $0.22$ & $95.62\pm 0.22$ & $96.93\pm 0.11$ & $96.61\pm 0.12$ & $97.26\pm 0.21$\\
    1 & 6& $0.13$ & $95.74\pm 0.17$ & $96.82\pm 0.06$ & $96.48\pm 0.13$ & $97.17\pm 0.16$\\
    \bottomrule\toprule
    \multicolumn{3}{c}{ScDCFNet}   &\multicolumn{2}{c}{SFashion $(28\times 28)$ accuracy (\%)}&\multicolumn{2}{c}{SFashion $(28 \times 28)+$ accuracy (\%)}\\
    \cmidrule(r){1-3} \cmidrule(r){4-5} \cmidrule(r){6-7}
    $K_\alpha$ & $K$ & \# Params  &$N_{\text{tr}} = 2,000$ & $N_{\text{tr}} = 5,000$ &  $N_{\text{tr}} = 2,000$ &$N_{\text{tr}} = 5,000$\\
    \midrule
    3 & 15 & $1.00$ & $\bm{81.32\pm 0.41}$ & $\bm{84.24\pm 0.35}$ & $\bm{82.42\pm 0.38}$ & $84.31\pm 0.30$\\
    3 & 12 & $0.80$ & $81.23\pm 0.37$ & $84.17\pm 0.30$ & $82.39\pm 0.46$ & $84.22\pm 0.40$\\
    3 & 10 & $0.67$ & $81.03\pm 0.38$ & $83.75\pm 0.23$ & $81.97\pm 0.50$ & $83.96\pm 0.42$\\
    3 & 6 & $0.40$ & $80.88\pm 0.27$ & $83.83\pm 0.11$ & $81.03\pm 0.22$ & $83.66\pm 0.21$\\
    2 & 15 & $0.67$ & $80.88\pm 0.31$ & $84.05\pm 0.40$ & $82.18\pm 0.34$ & $\bm{84.40\pm 0.30}$\\
    2 & 10 & $0.44$ & $80.65\pm 0.50$ & $83.89\pm 0.23$ & $81.86\pm 0.39$ & $83.85\pm 0.39$\\
    1 & 6 & $0.27$ & $80.82\pm 0.42$ & $83.44\pm 0.29$ & $81.46\pm 0.45$ & $83.62\pm 0.24$\\
    1 & 15 & $0.33$ & $80.74\pm 0.39$ & $84.09\pm 0.27$ & $82.33\pm 0.44$ & $84.17\pm 0.32$\\
    1 & 10 & $0.22$ & $80.89\pm 0.33$ & $83.65\pm 0.44$ & $81.87\pm 0.58$ & $83.66\pm 0.23$\\
    1 & 6 & $0.13$ & $80.60\pm 0.27$ & $83.61\pm 0.33$ & $81.40\pm 0.56$ & $83.57\pm 0.48$\\
    \bottomrule
  \end{tabular}
  \vspace{-.5em}
  \caption{\small Ablation study on the role of $K$ and $K_\alpha$ in the performance of ScDCFNet. The column ``\# Params'' stands for the number of parameters of the current model compared to that of the benchmark ScDCFNet used in Table~\ref{tab:acc_smnist} and Table~\ref{tab:acc_sfashion}, i.e., $K = 15$ and $K_\alpha=3$.}\label{tab:acc_ablation}
\end{table}

We next conduct an ablation study on the role of $K$ and $K_\alpha$, i.e., the numbers of the truncated  basis functions in space and scale, on the performance of ScDCFNet. The results of ScDCFNet on the SMNIST and SFashion data sets  with varying $K$ and $K_\alpha$ are shown in Table~\ref{tab:acc_ablation}. It can be observed that enlarging $K$ and $K_\alpha$, which increases the expressive power of the network at the cost of more trainable parameters (i.e., larger model size), indeed boosts the performance of ScDCFNet. However, the accuracy gradually plateaus around $K\approx 12$ and $K_\alpha\approx 3$. This is because high frequency filters suffer from aliasing effect after discretization, thus introducing a larger equivariance error. This phenomenon will be further explore in Section~\ref{sec:image_reconstruction}.

\subsubsection{STL-10}

\begin{table}[t]
  \centering
  \scriptsize
  \begin{tabular}{lc}
    \toprule  
    Models & Accuracy (\%) \\\midrule
    ResNet-16 & $81.98\pm 0.23$\\
    LSI-CNN & $81.79\pm 0.53$\\
    SI-CNN & $81.75\pm 0.14$ \\
    SS-CNN & $69.51\pm 1.27$ \\
    DSS &  $82.21\pm 0.47$ \\
    SESN & $83.89\pm 0.05$\\
    \midrule
    ScDCFNet & $\bm{84.90 \pm 0.36}$ \\
    \bottomrule
  \end{tabular}
  \caption{\small Test accuracy on the STL-10 data set.}\label{tab:acc_stl}
\end{table}

We next conduct the experiments on the STL-10 data set to examine the performance of various models on natural image classification. We use a ResNet by \cite{resnet} with 16 layers as the baseline, upon which all models are built while the number of trainable parameters is kept almost the same. Similarly, a max-pooling in the scale channel is performed only after the final residual block for equivariant models.  For SESN and ScDCFNet, the filter size in scale $L_\alpha$ is set to 2.

  Following the idea of \cite{sosnovik2019scale}, we augment the data set during training by applying 12 pixel zero-padding followed by random cropping. In addition, images are randomly flipped horizontally and Cutout by \cite{devries2017improved} with 1 hole of 32 pixels is used. All models are trained for 1000 epochs with a batch size of 128.  We use an SGD optimizer with Nesterov momentum set to 0.9 and weight decay being $5\times 10^{-4}$. The initial learning rate is set to 0.1 and scheduled to decrease tenfold after 300, 400, 600 and 800 epochs.

  We run three independent trials of the experiment, and report the mean~$\pm$~std of the test accuracy in Table~\ref{tab:acc_stl}. It is clear that ScDCFNet again achieves the best performance compared to other models, further demonstrating its advantage in multiscale image classification.

\subsection{Image Reconstruction}
\label{sec:image_reconstruction}

\begin{figure}[t]
    \centering
    \includegraphics[width=.9\textwidth]{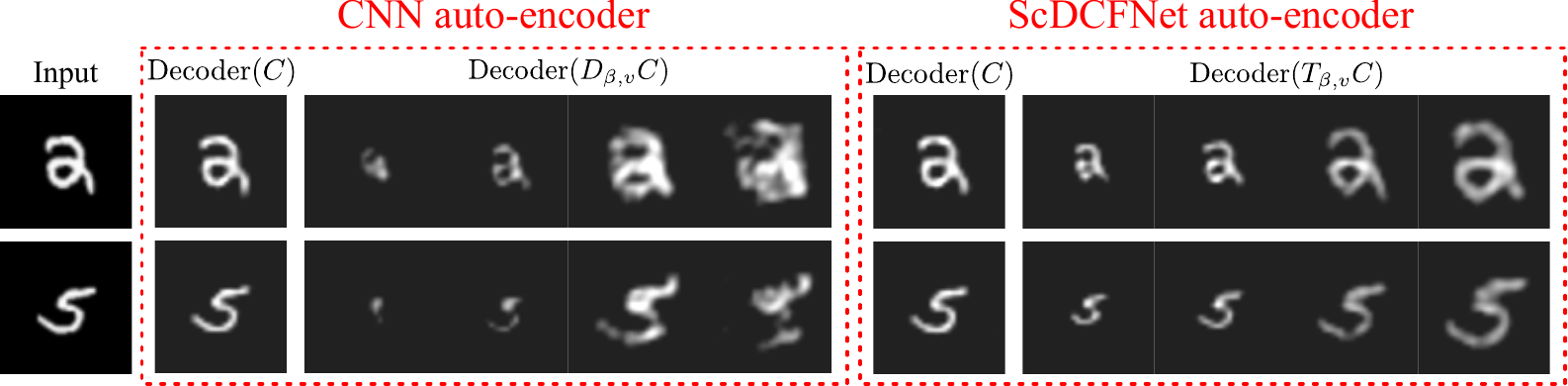}
    \caption{\small Reconstructing  rescaled versions of the original test image by manipulating its image code $C$ according to the group action \eqref{eq:group-action2}. The first two images on the left are the original inputs; $\text{Decoder}(C)$ denotes the reconstruction using the (unchanged) image code $C$; $\text{Decoder}(D_{\beta,v}C)$ and  $\text{Decoder}(T_{\beta,v}C)$ denote the reconstructions using the ``rescaled'' image codes $D_{\beta,v}C$ and $T_{\beta,v}C$ respectively according to \eqref{eq:group-action1} and \eqref{eq:group-action2}. Unlike the regular CNN auto-encoder,  the ScDCFNet auto-encoder manages to generate rescaled versions of the original input, suggesting that it successfully encodes the scale information directly into the representation.}
    \label{fig:auto}
\end{figure}

In the last experiment, we illustrate the ability of ScDCFNet to explicitly encode the input scale information into the representation. To achieve this, we train an ScDCFNet auto-encoder on the SMNIST data set with images resized to $56\times 56$ for better visualization. The encoder stacks two $\cst$-equivariant convolutional blocks with $2\times 2$  average-pooling, and the decoder contains a succession of two transposed convolutional blocks with $2\times 2$ upsampling. A regular CNN auto-encoder is also trained for comparison (see Table~\ref{tab:architecture_auto} in Appendix~\ref{sec:app_ex3} for the detailed architecture.)

Our goal is to demonstrate that the image code produced by the ScDCFNet auto-encoder contains the scale information of the input, i.e., by applying the group action $T_{\beta, v}$ \eqref{eq:group-action2} to the code $C$ of a test image before feeding it to the decoder, we can reconstruct rescaled versions of original input. This property can be visually verified in Figure~\ref{fig:auto}. In contrast,  a regular CNN auto-encoder fails to do so.

\begin{figure}[t]
    \centering
    \includegraphics[width=.7\textwidth]{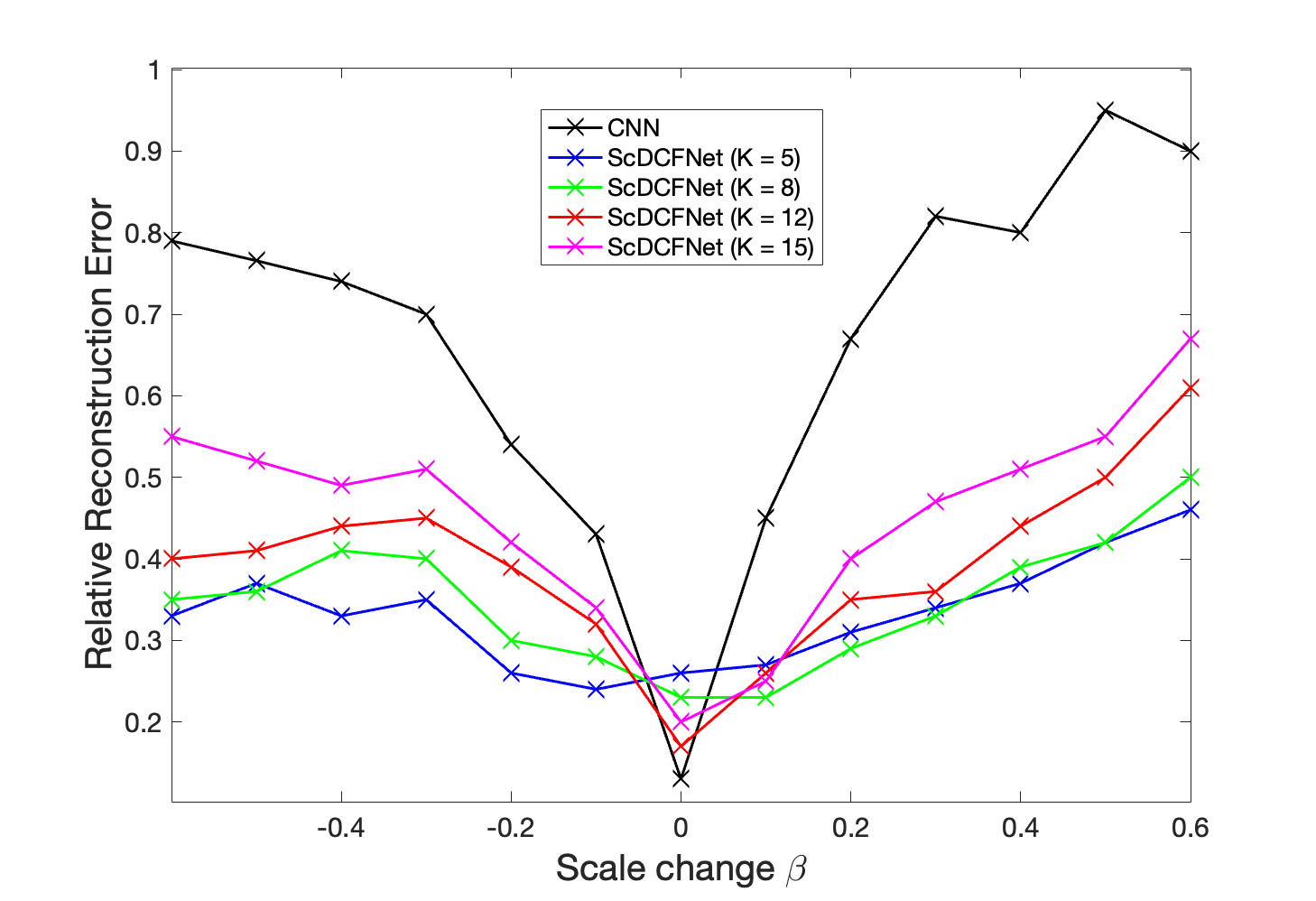}
    \caption{\small The relative error \eqref{eq:error_recon}  between the rescaled inputs and their reconstructions using CNN and ScDCFNet autoencoders at 13 different scales $2^{\beta}, \beta\in\{-0.6, -0.5, \cdots, 0.6\}$. The means of the reconstruction error over the test images of SMNIST are displayed.}
    \label{fig:im_recon}
\end{figure}

Finally, we quantitatively study the impact of basis truncation in ScDCFNet on the reconstruction error. In the experiment, we fix the number of basis functions in scale at $K_\alpha=3$, and train ScDCFNet auto-encoders with spatial basis functions  truncated to $K=5, 8, 12,$ and $15$. After training the networks, we calculate the relative $L^2$ distance between the rescaled original images $D_{\beta}[x^{(0)}]$ and their reconstructions at 13 different scales $2^{\beta}, \beta\in\{-0.6, -0.5, \cdots, 0.6\}$:
  \begin{align}
    \label{eq:error_recon}
  \text{Error} = \frac{\left\|\text{Decoder}\left\{T_{\beta}\left[\text{Encoder}\left(x^{(0)}\right)\right]\right\}-D_{\beta}\left[x^{(0)}\right]\right\|_{L^2}}{\left\|D_{\beta}\left[x^{(0)}\right]
  \right\|_{L^2}}.
  \end{align}
  To adjust for the contrast loss visible in Figure~\ref{fig:auto},  images are thresholded before calculating the error \eqref{eq:error_recon}. We display the mean of the reconstruction error over the test set of SMNIST across 13 different scales in Figure~\ref{fig:im_recon}. It can be observed that ScDCFNet autoencoders significantly outperform their CNN counterpart by having a smaller error when $\beta \not= 0$, i.e., reconstructing more accurately at a scale \textit{different} from the original image. Moreover, by reducing the number of basis functions $K$, the reconstruction error becomes smaller, demonstrating again our theoretical results Theorem~\ref{thm:regularity_final} and Theorem~\ref{thm:regularity_final_truncation} that basis truncation improves deformation robustness of the equivariant representation.

\section{Conclusion}
We propose, in this paper, a $\cst$-equivariant CNN with joint convolutions across the space $\R^2$ and the scaling group $\cs$, which we show to be both sufficient and necessary to achieve equivariance for the regular representation of the scaling-translation group. To reduce the computational cost and model complexity incurred by the joint convolutions, the convolutional filters supported on $\R^2\times\cs$ are decomposed under a separable basis across the two domains and truncated to only low-frequency components. Moreover, the truncated filter expansion leads also to improved deformation robustness of the equivariant representation, i.e., the model is still approximately equivariant even if the scaling transformation is imperfect. Experimental results suggest that ScDCFNet achieves improved performance in multiscale image classification with greater interpretability and reduced model size compared to regular CNN models.

For future work, we will study the application of ScDCFNet in other more complicated vision tasks, such as object detection/localization and pose estimation,  where it is beneficial to directly encode the input scale information into the deep representation. We will explore other efficient implementation of the model, e.g., using filter-bank type of techniques to compute convolutions with multiscale spatial filters, to  further reduce the computational cost.


\section*{Acknowledgement}
The research of WZ is partially supported by NSF under DMS-2052525 and DMS-2140982. QQ is partially supported by NSF DMS-1737744. The research of RC is supported in part by the Air Force Office of Scientific Research through award FA 9550-20-1-0266. GS is partially supported by NSF, NGA, and ONR. XC is partially supported by NSF (DMS-1820827) and the Alfred P. Sloan Foundation.


\appendix

\section{Proofs}

\subsection{Proof of Theorem~\ref{thm:equivariance}}
\begin{proof}[Proof of Theorem~\ref{thm:equivariance}]
  We note first that \eqref{eq:scale-equivariant} holds true if and only if the following being valid for all $l\ge 1$,
  \begin{align}
    \label{eq:scale-equivariant-adj}
    T_{\beta, v}x^{(l)}[x^{(l-1)}] = x^{(l)}[T_{\beta, v} x^{(l-1)}],
  \end{align}
where $T_{\beta, v}x^{(0)}$ is understood as $D_{\beta, v}x^{(0)}$. We also note that the layerwise operations of a general feedforward neural network with an extra index $\alpha\in \cs$ can be written as
\begin{align}
  \label{eq:1st-layer-cnn}
  x^{(1)}[x^{(0)}](u,\alpha, \lambda) = \sigma\left( \sum_{\lambda'}\int_{\R^2} x^{(0)}(u+u', \lambda')W^{(1)}(u',\lambda',u,\alpha,\lambda) du'+b^{(1)}(\lambda)\right),
\end{align}
and, for $l>1$,
\begin{align}
  \nonumber
  x^{(l)}[x^{(l-1)}](u,\alpha, \lambda) = &\sigma\left( \sum_{\lambda'}\int_{\R^2}\int_{\R} x^{(l-1)}(u+u', \alpha+\alpha', \lambda')\right.\\   \label{eq:lth-layer-cnn}
  &\quad\quad\quad\quad\quad\quad \left.W^{(l)}(u', \alpha',\lambda',u,\alpha,\lambda)d\alpha' du'+b^{(l)}(\lambda)\right).
\end{align}

To prove the sufficient part: when $l=1$, \eqref{eq:group-action1}, \eqref{eq:group-action2}, and \eqref{eq:1st-layer} lead to
\begin{align}
  &T_{\beta, v}x^{(1)}[x^{(0)}](u,\alpha,\lambda) = x^{(1)}[x^{(0)}]\left(2^{-\beta}(u-v), \alpha-\beta, \lambda \right) \\
  = & \sigma\left( \sum_{\lambda'}\int x^{(0)}\left( 2^{-\beta}(u-v)+u',\lambda' \right)W^{(1)}_{\lambda',\lambda}\left(2^{-(\alpha-\beta)}u'\right)2^{-2(\alpha-\beta)}du'+b^{(1)}(\lambda) \right)\\
  = & \sigma\left( \sum_{\lambda'}\int x^{(0)}\left( 2^{-\beta}(u-v+\tilde{u}),\lambda' \right)W^{(1)}_{\lambda',\lambda}\left(2^{-\alpha}\tilde{u}\right)2^{-2\alpha}d\tilde{u}+b^{(1)}(\lambda) \right),
\end{align}
and
\begin{align}\nonumber
  &x^{(1)}[D_{\beta, v}x^{(0)}](u,\alpha,\lambda)\\
  = & \sigma\left( \sum_{\lambda'}\int_{\R^2} D_{\beta, v}x^{(0)}(u+u', \lambda') W_{\lambda', \lambda}^{(1)}\left(2^{-\alpha}u'\right)2^{-2\alpha}du'+b^{(1)}(\lambda)\right)\\
  = & \sigma\left( \sum_{\lambda'}\int_{\R^2}x^{(0)}(2^{-\beta}\left(u+u'-v), \lambda'\right) W_{\lambda', \lambda}^{(1)}\left(2^{-\alpha}u'\right)2^{-2\alpha}du'+b^{(1)}(\lambda)\right).
\end{align}
Hence $T_{\beta, v}x^{(1)}[x^{(0)}] = x^{(1)}[D_{\beta, v}x^{(0)}]$.

When $l>1$, we have
\begin{align}
  &T_{\beta, v}x^{(l)}[x^{(l-1)}](u,\alpha,\lambda) = x^{(l)}[x^{(l-1)}]\left(2^{-\beta}(u-v), \alpha-\beta, \lambda \right) \\\nonumber
  = & \sigma\left( \sum_{\lambda'}\int\int x^{(l-1)}\left( 2^{-\beta}(u-v)+u', \alpha-\beta+\alpha',\lambda' \right)\right.\cdot\\
  &\quad\quad\quad\quad\quad\left.W^{(l)}_{\lambda',\lambda}\left(2^{-(\alpha-\beta)}u', \alpha'\right)2^{-2(\alpha-\beta)}du'd\alpha'+b^{(l)}(\lambda) \right)\\\nonumber
  = & \sigma\left( \sum_{\lambda'}\int\int x^{(l-1)}\left( 2^{-\beta}(u-v+\tilde{u}), \alpha-\beta+\alpha',\lambda' \right)\right.\cdot\\
  & \quad\quad\quad\quad\quad\left.W^{(l)}_{\lambda',\lambda}\left(2^{-\alpha}\tilde{u}, \alpha'\right)2^{-2\alpha}d\tilde{u}d\alpha'+b^{(l)}(\lambda) \right)
\end{align}
and
\begin{align}\nonumber
  &x^{(l)}[T_{\beta,v}x^{(l-1)}](u,\alpha,\lambda)\\
  = & \sigma\left( \sum_{\lambda'}\int\int T_{\beta,v}x^{(l-1)}(u+u', \alpha+\alpha', \lambda') W_{\lambda', \lambda}^{(l)}\left(2^{-\alpha}u', \alpha'\right)2^{-2\alpha}d\alpha' du'+b^{(l)}(\lambda)\right)\\\nonumber
  = & \sigma\left( \sum_{\lambda'}\int\int x^{(l-1)}(2^{-\beta}(u+u'-v), \alpha+\alpha'-\beta, \lambda') \right.\cdot\\
  &\quad\quad\quad\quad\quad\quad\quad\left.W_{\lambda', \lambda}^{(l)}\left(2^{-\alpha}u', \alpha'\right)2^{-2\alpha}d\alpha' du'+b^{(l)}(\lambda)\right)
\end{align}
Therefore $T_{\beta, v}x^{(l)}[x^{(l-1)}] = x^{(l)}[T_{\beta, v} x^{(l-1)}], ~\forall l>1$.

To prove the necessary part: when $l=1$, we have
\begin{align}
  &T_{\beta, v}x^{(1)}[x^{(0)}](u,\alpha,\lambda) = x^{(1)}[x^{(0)}]\left(2^{-\beta}(u-v), \alpha-\beta, \lambda \right) \\\nonumber
  = & \sigma\left( \sum_{\lambda'}\int x^{(0)}\left( 2^{-\beta}(u-v)+u',\lambda' \right)\right.\cdot\\
  & \quad\quad\quad\quad\left.W^{(1)}\left(u',\lambda',2^{-\beta}(u-v), \alpha-\beta, \lambda\right)du'+b^{(1)}(\lambda) \right)\\ \nonumber
 = & \sigma\left( \sum_{\lambda'}\int x^{(0)}\left( 2^{-\beta}(u+u'-v),\lambda' \right)\right.\cdot\\
  & \quad\quad\quad\quad\left.W^{(1)}\left(2^{-\beta}u',\lambda',2^{-\beta}(u-v), \alpha-\beta, \lambda\right)2^{-2\beta}du'+b^{(1)}(\lambda) \right),
\end{align}
and
\begin{align}\nonumber
  & x^{(1)}[D_{\beta, v}x^{(0)}](u,\alpha,\lambda)\\
  = & \sigma\left(\sum_{\lambda'}\int D_{\beta, v}x^{(0)}\left( u+u', \lambda'\right)W^{(1)}\left(u',\lambda',u,\alpha,\lambda \right)du'+b^{(1)}(\lambda) \right) \\
  = & \sigma\left(\sum_{\lambda'}\int x^{(0)}\left( 2^{-\beta}(u+u'-v), \lambda'\right)W^{(1)}\left(u',\lambda',u,\alpha,\lambda \right)du'+b^{(1)}(\lambda) \right)
\end{align}
Hence for \eqref{eq:scale-equivariant-adj} to hold when $l=1$, we need
\begin{align}
  \label{eq:necessary_l=1}
  W^{(1)}\left(u',\lambda',u,\alpha,\lambda\right) = W^{(1)}\left(2^{-\beta}u',\lambda',2^{-\beta}(u-v),\alpha-\beta,\lambda\right)2^{-2\beta}
\end{align}
to hold true for all $u,\alpha,\lambda,u',\lambda',v,\beta$. Keeping $u,\alpha,\lambda,u',\lambda', \beta$ fixed while changing $v$ in \eqref{eq:necessary_l=1}, we obtain that $W^{(1)}(u',\lambda',u,\alpha,\lambda)$ does not depend on the third variable $u$. Thus $W^{(1)}\left(u',\lambda',u,\alpha,\lambda\right) = W^{(1)}\left(u',\lambda',0,\alpha,\lambda\right),~\forall u$. Define $W_{\lambda,\lambda}^{(1)}(u')$ as
\begin{align}
  W^{(1)}_{\lambda',\lambda}(u') \coloneqq W^{(1)}\left(u',\lambda',0,0,\lambda\right).
\end{align}
Then, for any given $u',\lambda',u,\alpha,\lambda$, setting $\beta = \alpha$ in \eqref{eq:necessary_l=1} leads to
\begin{align}
  &W^{(1)}\left(u',\lambda',u,\alpha,\lambda\right) = W^{(1)}\left(2^{-\alpha}u',\lambda',2^{-\alpha}(u-v),0,\lambda\right)2^{-2\alpha}\\
  =& W^{(1)}\left(2^{-\alpha}u',\lambda',0,0,\lambda\right)2^{-2\alpha}  = W_{\lambda',\lambda}^{(1)}\left(2^{-\alpha}u'\right)2^{-2\alpha}.
\end{align}
Hence \eqref{eq:1st-layer-cnn} can be written as \eqref{eq:1st-layer}.

For $l>1$, a similar argument leads to
\begin{align}
  \label{eq:necessary_l>1}
  W^{(l)}\left(u', \alpha',\lambda',u,\alpha,\lambda\right) = W^{(l)}\left(2^{-\beta}u', \alpha',\lambda',2^{-\beta}(u-v),\alpha-\beta,\lambda\right)2^{-2\beta}
\end{align}
for all $u,\alpha,\lambda,u',\alpha',\lambda',v,\beta$. Again, keeping $u,\alpha,\lambda,u', \alpha',\lambda', \beta$ fixed while changing $v$ in \eqref{eq:necessary_l>1} leads us to the conclusion that $W^{(l)}(u', \alpha',\lambda',u,\alpha,\lambda)$ does not depend on the fourth variable $u$. Define
\begin{align}
  W^{(l)}_{\lambda',\lambda}(u', \alpha') \coloneqq W^{(l)}\left(u', \alpha',\lambda',0,0,\lambda\right).
\end{align}
After setting $\beta = \alpha$ in \eqref{eq:necessary_l>1}, for any given $u', \alpha',\lambda',u,\alpha,\lambda$, we have
\begin{align}
  &W^{(l)}\left(u', \alpha',\lambda',u,\alpha,\lambda\right) = W^{(l)}\left(2^{-\alpha}u', \alpha',\lambda',2^{-\alpha}(u-v),0,\lambda\right)2^{-2\alpha}\\
  =& W^{(l)}\left(2^{-\alpha}u', \alpha',\lambda',0,0,\lambda\right)2^{-2\alpha}  = W_{\lambda',\lambda}^{(l)}\left(2^{-\alpha}u'\right)2^{-2\alpha}.
\end{align}
This concludes the proof of the Theorem.
\end{proof}

\subsection{Proof of Proposition~\ref{prop:computational-cost}}
\begin{proof}[Proof of Theorem~\ref{prop:computational-cost}]
  In a regular $\cst$-equivariant CNN, the $l$-th convolutional layer \eqref{eq:lth-layer} is computed as follows:
  \begin{align}
    \label{eq:regular-step1}
    y(u,\alpha,\alpha',\lambda, \lambda') &= \int_{\R^2} x^{(l-1)}(u+u', \alpha+\alpha', \lambda')W_{\lambda',\lambda}^{(l)}\left(2^{-\alpha}u', \alpha' \right)2^{-2\alpha}du',\\
    \label{eq:regular-step2}
    z(u,\alpha,\lambda,\lambda') & = \int_\R y(u,\alpha,\alpha',\lambda,\lambda')d\alpha',\\
    \label{eq:regular-step3}
    x^{(l)}(u,\alpha,\lambda) &= \sigma\left(\sum_{\lambda'=1}^{M_{l-1}}z(u,\alpha,\lambda,\lambda') + b^{(l)}(\lambda)\right).
  \end{align}
The spatial convolutions in \eqref{eq:regular-step1} take $2HWL^2N_s L_\alpha M_l M_{l-1}$ flops (there are $N_s L_\alpha M_l M_{l-1}$ convolutions in $u$, each taking $2HWL^2$ flops.) The summation over $\alpha'$ in \eqref{eq:regular-step2} requires $L_\alpha N_s HW M_l M_{l-1}$ flops. The summation over $\lambda'$, adding the bias, and applying the nonlinear activation in \eqref{eq:regular-step3} requires an additional $HWN_s M_l(2+M_{l-1})$ flops. Thus the total number of floating point computations in a forward pass through the $l$-th layer of a regular $\cst$-equivariant CNN is
\begin{align}
  \label{eq:flops-regular}
  HWN_s M_l(2L^2L_\alpha M_{l-1}+L_\alpha M_{l-1} + M_{l-1}+2).
\end{align}

On the other hand, in an ScDCFNet with separable basis truncation up to $KK_{\alpha}$ leading coefficients, \eqref{eq:lth-layer} can be computed via the following steps:
\begin{align}
  \label{eq:sdcfnet-step1}
  y(u,\alpha,\lambda',m) &=  \int_{\R}x^{(l-1)}(u,\alpha+\alpha', \lambda')\varphi_m(\alpha')d\alpha'\\
  \label{eq:sdcfnet-step2}
  z(u,\alpha,\lambda',k,m) & = \int_{\R^2}y(u+u',\alpha,\lambda',m)\psi_{j_l,k}(2^{-\alpha}u')2^{-2\alpha}du'\\
  \label{eq:sdcfnet-step3}
  x^{(l)}(u,\alpha,\lambda) &= \sigma\left(\sum_{m=1}^{K_\alpha}\sum_{k=1}^K\sum_{\lambda'=1}^{M_{l-1}}z(u,\alpha,\lambda',k,m)a_{\lambda',\lambda}^{(l)}(k,m) + b^{(l)}(\lambda)\right).  
\end{align}
The convolutions in $\alpha$ \eqref{eq:sdcfnet-step1} require $2L_{\alpha}N_{s}HWK_{\alpha}M_{l-1}$ flops (there are $HWK_{\alpha}M_{l-1}$ convolutions in $\alpha$, each taking $2L_{\alpha}N_{s}$ flops.) The spatial convolutions in \eqref{eq:sdcfnet-step2} take $2HWL^2N_{s}M_{l-1}K_{\alpha}K$ flops ($N_{s}M_{l-1}K_{\alpha}K$ convolutions in $u$, each taking $2HWL^2$ flops.) The last step \eqref{eq:sdcfnet-step3} requires an additional $2HWN_s M_l(1+K K_\alpha M_{l-1})$ flops. Hence the total number of floating point computation for an ScDCFNet is 
\begin{align}
  \label{eq:flops-sdcfnet}
  2HWN_{s}(KK_\alpha M_{l-1}M_l + M_l + L^2M_{l-1}K_\alpha K + L_\alpha K_\alpha M_{l-1}).
\end{align}

In particular, when $M_l \gg L^2, L_\alpha$, the dominating terms in \eqref{eq:flops-regular} and \eqref{eq:flops-sdcfnet} are, respectively, $2HWN_s M_l M_{l-1}L^2L_\alpha$ and $2HWN_{s} M_{l-1}M_lKK_\alpha$. Thus the computational cost in an ScDCFNet has been reduced to a factor of $\frac{K K_\alpha}{L^2 L_\alpha}$.
\end{proof}


\subsection{Proof of Proposition~\ref{prop:nonexpansive}}

Before proving Proposition~\ref{prop:nonexpansive}, we need the following two lemmas.
\begin{lemma}
  \label{lemma:FB-bounds-everything}
  Suppose that $\left\{\psi_k\right\}_k$ are the the spatial bases defined in \eqref{eq:def-fb}, and
  \begin{align}
    F(u) = \sum_k a(k)\psi_{j,k}(u) = \sum_{k}a(k)2^{-2j}\psi_k(2^{-j}u)
  \end{align}
  is a smooth function on $[-2^j, 2^j]^2$, then
  \begin{align}
    \label{eq:FB-bounds-everything}
    \int \left| F(u)\right|du, ~\int \left|u\right|\left|\nabla F(u)\right|du, ~2^j\int \left|\nabla F(u)\right|du \le \pi\|a\|_{\mu} = \pi \left(\sum_k \mu_k \cdot a(k)^2\right)^{1/2}.
  \end{align}
\end{lemma}
This is essentially Lemma 3.5 and Proposition 3.6 in \cite{dcfnet} after rescaling $u$. The only difference is that the basis functions $\psi_k(u)$ defined in \eqref{eq:def-fb} are eigenfunctions of Dirichlet Laplacian on $[-1, 1]^2$ instead of the unit disk. The key elements of the proof, i.e., the orthogonality of $\psi_k$ and the Poincar\'e inequality, still apply. Lemma~\ref{lemma:FB-bounds-everything} easily leads to the following lemma.

\begin{lemma}
  \label{lemma:Al-bounds-everything}
  Let $a_{\lambda',\lambda}^{(l)}(k,m)$ be the coefficients of the filter $W_{\lambda',\lambda}^{(l)}(u,\alpha)$ under the joint bases $\left\{\psi_k\right\}_k$ and $\left\{\varphi_m\right\}_m$ defined in \eqref{eq:joint-bases}, and define $W_{\lambda',\lambda,m}^{(l)}(u)$ as
  \begin{align}
    \label{eq:W_lambda'_lambda_m}
    W_{\lambda',\lambda,m}(u)\coloneqq \sum_k a_{\lambda',\lambda}^{(l)}(k,m)\psi_{j_l,k}(u).
  \end{align}
  We have, for all $l>1$,
  \begin{align}
    B^{(1)}_{\lambda',\lambda}, C^{(1)}_{\lambda',\lambda}, 2^{j_1}D^{(1)}_{\lambda',\lambda} \le \pi\|a^{(1)}_{\lambda',\lambda}\|_{\mu},  ~~ B^{(l)}_{\lambda',\lambda,m}, C^{(l)}_{\lambda',\lambda,m}, 2^{j_l}D^{(l)}_{\lambda',\lambda,m} \le \pi\|a^{(l)}_{\lambda',\lambda}(\cdot, m)\|_{\mu},
  \end{align}
where
\begin{align}
  \label{eq:BCD_lambda}
  \left\{
  \begin{aligned}
    &B_{\lambda',\lambda}^{(1)}\coloneqq \int\left|W_{\lambda',\lambda}^{(1)}(u)\right|du, \quad &&B_{\lambda',\lambda,m}^{(l)}\coloneqq \int\left|W_{\lambda',\lambda,m}^{(l)}(u)\right|du, \quad l>1,\\
    &C_{\lambda',\lambda}^{(1)}\coloneqq \int\left|u\right| \left|\nabla_uW_{\lambda',\lambda}^{(1)}(u)\right|du, \quad &&C_{\lambda',\lambda,m}^{(l)}\coloneqq \int\left|u\right| \left|\nabla_uW_{\lambda',\lambda,m}^{(l)}(u)\right|du, \quad l>1,\\
    &D_{\lambda',\lambda}^{(1)}\coloneqq \int \left|\nabla_uW_{\lambda',\lambda}^{(1)}(u)\right|du, \quad &&D_{\lambda',\lambda,m}^{(l)}\coloneqq \int \left|\nabla_uW_{\lambda',\lambda,m}^{(l)}(u)\right|du, \quad l>1.\\
  \end{aligned}\right.
\end{align}
We thus have
\begin{align}
  B_l, C_l, 2^{j_l}D_l \le A_l,
\end{align}
where
\begin{align}
  \label{eq:BCD1}
  \begin{aligned}
    &B_1 \coloneqq \max \left\{\sup_{\lambda}\sum_{\lambda'=1}^{M_0}B_{\lambda',\lambda}^{(1)}, ~\frac{M_0}{M_1}\sup_{\lambda'}\sum_{\lambda=1}^{M_1}B_{\lambda',\lambda}^{(1)}\right\},\\
    &C_1 \coloneqq \max \left\{\sup_{\lambda}\sum_{\lambda'=1}^{M_0}C_{\lambda',\lambda}^{(1)}, ~\frac{M_0}{M_1}\sup_{\lambda'}\sum_{\lambda=1}^{M_1}C_{\lambda',\lambda}^{(1)}\right\},\\
    &D_1 \coloneqq \max \left\{\sup_{\lambda}\sum_{\lambda'=1}^{M_0}D_{\lambda',\lambda}^{(1)}, ~\frac{M_0}{M_1}\sup_{\lambda'}\sum_{\lambda=1}^{M_1}D_{\lambda',\lambda}^{(1)}\right\},
  \end{aligned}
\end{align}
and, for $l>1$,
\begin{align}
  \label{eq:BCDl}
  \begin{aligned}
    &B_l \coloneqq \max \left\{\sup_{\lambda}\sum_{\lambda'=1}^{M_{l-1}}\sum_m B_{\lambda',\lambda,m}^{(l)}, ~\frac{2M_{l-1}}{M_l}\sum_m B_{l,m}\right\}, \quad B_{l,m}\coloneqq\sup_{\lambda'}\sum_{\lambda=1}^{M_l}B_{\lambda',\lambda,m}^{(l)},\\
    &C_l \coloneqq \max \left\{\sup_{\lambda}\sum_{\lambda'=1}^{M_{l-1}}\sum_m C_{\lambda',\lambda,m}^{(l)}, ~\frac{2M_{l-1}}{M_l}\sum_m C_{l,m}\right\}, \quad C_{l,m}\coloneqq\sup_{\lambda'}\sum_{\lambda=1}^{M_l}C_{\lambda',\lambda,m}^{(l)},\\
    &D_l \coloneqq \max \left\{\sup_{\lambda}\sum_{\lambda'=1}^{M_{l-1}}\sum_m D_{\lambda',\lambda,m}^{(l)}, ~\frac{2M_{l-1}}{M_l}\sum_m D_{l,m}\right\}, \quad D_{l,m}\coloneqq\sup_{\lambda'}\sum_{\lambda=1}^{M_l}D_{\lambda',\lambda,m}^{(l)}.
  \end{aligned}
\end{align}
In particular, (A2) implies that $B_l, C_l, 2^{j_l}D_l \le 1, ~\forall l$.
\end{lemma}

\begin{proof}[Proof of Proposition~\ref{prop:nonexpansive}]
   To simplify the notation, we omit $(l)$ in $W_{\lambda',\lambda}^{(l)}$, $W_{\lambda',\lambda,m}^{(l)}$, and $b^{(l)}$, and let $M = M_l$, $M' = M_{l-1}$. The proof of (a) for the case $l=1$ is similar to Proposition~3.1(a) of \cite{dcfnet} after noticing the fact that
  \begin{align}
    \label{eq:change-of-variable}
    \int_{\R^2}\left| W(2^{-\alpha}u)\right|2^{-2\alpha}du  = \int_{\R^2}\left| W(u)\right|du,
  \end{align}
and we include it here for completeness. From the definition of $B_1$ in \eqref{eq:BCD1}, we have
\begin{align}
  \sup_{\lambda}\sum_{\lambda'}B_{\lambda',\lambda}^{(1)} \le B_1, \quad \text{and}~~ \sup_{\lambda'}\sum_{\lambda}B_{\lambda',\lambda}^{(1)} \le B_1\frac{M}{M'}.
\end{align}
Thus, given two arbitrary functions $x_1$ and $x_2$, we have
\begin{align}
  &\left|\left(x^{(1)}[x_1]-x^{(1)}[x_2]\right)(u,\alpha,\lambda)\right|^2\\
  = & \left|\sigma\left( \sum_{\lambda'}\int x_1(u+u', \lambda') W_{\lambda', \lambda}\left(2^{-\alpha}u'\right)2^{-2\alpha}du'+b(\lambda)\right)\right.\\
  & \left. - \sigma\left( \sum_{\lambda'}\int x_2(u+u', \lambda') W_{\lambda', \lambda}\left(2^{-\alpha}u'\right)2^{-2\alpha}du'+b(\lambda)\right)\right|^2\\\nonumber
  \le &\left| \sum_{\lambda'}\int x_1(u+u', \lambda') W_{\lambda', \lambda}\left(2^{-\alpha}u'\right)2^{-2\alpha}du' \right.\\
  &\quad \quad \left.- \sum_{\lambda'}\int x_2(u+u', \lambda') W_{\lambda', \lambda}\left(2^{-\alpha}u'\right)2^{-2\alpha}du' \right|^2\\
  = & \left| \sum_{\lambda'}\int (x_1 - x_2)(u+u', \lambda') W_{\lambda', \lambda}\left(2^{-\alpha}u'\right)2^{-2\alpha}du' \right|^2\\
  \le & \left( \sum_{\lambda'}\int \left| (x_1-x_2)(u+u',\lambda')\right|^2\left|W_{\lambda',\lambda}(2^{-\alpha}u')\right|2^{-2\alpha}du' \right) \sum_{\lambda'}\int \left|W_{\lambda',\lambda}(2^{-\alpha}u')\right|2^{-2\alpha}du'\\
  = & \left( \sum_{\lambda'}\int \left| (x_1-x_2)(u+u',\lambda')\right|^2\left|W_{\lambda',\lambda}(2^{-\alpha}u')\right|2^{-2\alpha}du' \right) \left( \sum_{\lambda'}B_{\lambda',\lambda}^{(1)} \right)\\
  \le & B_1\sum_{\lambda'}\int \left| (x_1-x_2)(\tilde{u},\lambda')\right|^2\left|W_{\lambda',\lambda}(2^{-\alpha}\left(\tilde{u}-u)\right)\right|2^{-2\alpha}d\tilde{u}
\end{align}
Therefore, for any $\alpha$,
\begin{align}
  \nonumber
  & \sum_{\lambda}\int \left|\left(x^{(1)}[x_1]-x^{(1)}[x_2]\right)(u,\alpha,\lambda)\right|^2 du \\
  \le & \sum_{\lambda}\int B_1\sum_{\lambda'}\int \left| (x_1-x_2)(\tilde{u},\lambda')\right|^2\left|W_{\lambda',\lambda}(2^{-\alpha}\left(\tilde{u}-u)\right)\right|2^{-2\alpha}d\tilde{u} du\\
  = & B_1\sum_{\lambda'}\int \left| (x_1-x_2)(\tilde{u},\lambda')\right|^2  \left( \sum_{\lambda}\int \left|W_{\lambda',\lambda}(2^{-\alpha}\left(\tilde{u}-u)\right)\right|2^{-2\alpha} du\right)d\tilde{u}\\
  = & B_1\sum_{\lambda'}\int \left| (x_1-x_2)(\tilde{u},\lambda')\right|^2 \left( \sum_{\lambda}B_{\lambda',\lambda}^{(1)}\right)d\tilde{u}\\
  \le & B_1^2\frac{M}{M'} \sum_{\lambda'}\int \left| (x_1-x_2)(\tilde{u},\lambda')\right|^2 d\tilde{u}\\
  = & B_1^2 M \|x_1 - x_2\|^2\\ 
  \le & M \|x_1 - x_2\|^2,
\end{align}
where the last inequality makes use of the fact that $B_1\le A_1\le 1$ under (A2) (Lemma~\ref{lemma:Al-bounds-everything}.) Therefore
\begin{align}
  \|x^{(1)}[x_1] - x^{(1)}[x_2] \|^2 & = \sup_{\alpha}\frac{1}{M}\sum_{\lambda}\int \left|\left(x^{(1)}[x_1]-x^{(1)}[x_2]\right)(u,\alpha,\lambda)\right|^2 du\\
                                     & \le \|x_1-x_2\|^2.
\end{align}
This concludes the proof of (a) for the case $l=1$. To prove the case for any $l>1$, we first recall from \eqref{eq:BCDl} that
\begin{align}
  \sup_{\lambda}\sum_{\lambda'}\sum_m B_{\lambda',\lambda,m}^{(l)} \le B_l, \quad \text{and}~~ \sum_m B_{l,m}\le B_l\frac{M}{2M'}, \quad\text{where}~~ B_{l,m}=\sup_{\lambda'}\sum_{\lambda}B_{\lambda',\lambda,m}^{(l)}.
\end{align}
Thus, for two arbitrary functions $x_1$ and $x_2$, we have
\begin{align}
  & \left|\left(x^{(l)}[x_1] - x^{(l)}[x_2]\right)(u,\alpha,\lambda)\right|^2\\ \nonumber
  = & \left|\sigma\left( \sum_{\lambda'}\int_{\R^2}\int_{\R} x_1(u+u', \alpha+\alpha', \lambda') W_{\lambda', \lambda}\left(2^{-\alpha}u', \alpha'\right)2^{-2\alpha}d\alpha' du'+b(\lambda)\right) \right.\\
  & \left.- \sigma\left( \sum_{\lambda'}\int_{\R^2}\int_{\R} x_2(u+u', \alpha+\alpha', \lambda') W_{\lambda', \lambda}\left(2^{-\alpha}u', \alpha'\right)2^{-2\alpha}d\alpha' du'+b(\lambda)\right) \right|^2\\
  \le & \left|  \sum_{\lambda'}\int_{\R^2}\int_{\R} (x_1-x_2)(u+u', \alpha+\alpha', \lambda') W_{\lambda', \lambda}\left(2^{-\alpha}u', \alpha'\right)2^{-2\alpha}d\alpha' du' \right|^2\\
  = & \left|  \sum_{\lambda'}\int_{\R^2}\int_{\R} (x_1-x_2)(u+u', \alpha+\alpha', \lambda')2^{-2\alpha}\sum_m W_{\lambda', \lambda,m}\left(2^{-\alpha}u'\right)\varphi_m(\alpha')  d\alpha' du' \right|^2\\
  = &\left| \sum_{\lambda'}\sum_{m}\int_{\R^2}G_m(u+u', \alpha, \lambda') 2^{-2\alpha}W_{\lambda',\lambda,m}(2^{-\alpha}u')du'\right|^2\\\nonumber
  \le &\left(\sum_{\lambda'}\sum_m \int_{\R^2}\left|G_m(u+u',\alpha,\lambda')\right|^2\left|W_{\lambda',\lambda,m}(2^{-\alpha}u')\right|2^{-2\alpha} du'\right)\cdot\\
  &\quad\left(\sum_{\lambda'}\sum_m \int_{\R^2}\left|W_{\lambda',\lambda,m}(2^{-\alpha}u')\right|2^{-2\alpha} du'\right)\\
  = &\left(\sum_{\lambda'}\sum_m \int_{\R^2}\left|G_m(\tilde{u},\alpha,\lambda')\right|^2\left|W_{\lambda',\lambda,m}(2^{-\alpha}(\tilde{u}-u))\right|2^{-2\alpha} d\tilde{u}\right)\left(\sum_{\lambda'}\sum_m B_{\lambda',\lambda,m}^{(l)}\right)\\
  \le & B_l\sum_{\lambda'}\sum_m \int_{\R^2}\left|G_m(\tilde{u},\alpha,\lambda')\right|^2\left|W_{\lambda',\lambda,m}(2^{-\alpha}(\tilde{u}-u))\right|2^{-2\alpha} d\tilde{u},
\end{align}
where
\begin{align}
  \label{eq:Gm_def}
  G_m(u,\alpha,\lambda') \coloneqq \int_\R (x_1-x_2)(u,\alpha+\alpha', \lambda')\varphi_m(\alpha')d\alpha'.
\end{align}
We claim (to be proved later in Lemma~\ref{lemma:Gm_bound}) that
\begin{align}
  M'\|G_m\|^2 = \sup_\alpha \sum_{\lambda'}\int_{\R^2}\left| G_m(u,\alpha,\lambda')\right|^2du \le 2M'\|x_1-x_2\|^2, \quad \forall m.
\end{align}
Thus, for any $\alpha$,
\begin{align}
  & \sum_{\lambda}\int_{\R^2} \left|\left(x^{(l)}[x_1] - x^{(l)}[x_2]\right)(u,\alpha,\lambda)\right|^2 du\\
  \le & \sum_{\lambda}\int_{\R^2}B_l\sum_{\lambda'}\sum_m \int_{\R^2}\left|G_m(\tilde{u},\alpha,\lambda')\right|^2\left|W_{\lambda',\lambda,m}(2^{-\alpha}(\tilde{u}-u))\right|2^{-2\alpha} d\tilde{u} du\\
  = & B_l\sum_{\lambda'}\sum_m\int_{\R^2}\left|G_m(\tilde{u},\alpha,\lambda') \right|^2 \left(\sum_{\lambda}\int_{\R^2}\left| W_{\lambda',\lambda,m}(2^{-\alpha}(\tilde{u}-u)) \right|2^{-2\alpha} du \right)d\tilde{u} \\
  = & B_l\sum_{\lambda'}\sum_m\int_{\R^2}\left|G_m(\tilde{u},\alpha,\lambda') \right|^2\left(\sum_{\lambda}B_{\lambda',\lambda,m}^{(l)} \right) d\tilde{u}\\
  \le & B_l\sum_{\lambda'}\sum_m\int_{\R^2}\left|G_m(\tilde{u},\alpha,\lambda') \right|^2 B_{l,m} d\tilde{u}\\
  = & B_l\sum_m\left( \sum_{\lambda'}\int_{\R^2}\left|G_m(\tilde{u},\alpha,\lambda') \right|^2 d\tilde{u}\right)B_{l,m}\\
  \le & B_l\cdot 2M'\|x_1-x_2\|^2  \sum_m B_{l,m}\\
  \le & B_l^2\cdot 2M'\|x_1-x_2\|^2 \frac{M}{2M'} \le M\|x_1-x_2\|^2.
\end{align}
Therefore
\begin{align}
  \|x^{(l)}[x_1] - x^{(l)}[x_2]\|^2 & = \sup_\alpha \frac{1}{M}\sum_{\lambda}\int_{\R^2} \left|\left(x^{(l)}[x_1] - x^{(l)}[x_2]\right)(u,\alpha,\lambda)\right|^2 du\\
                                    & \le \|x_1-x_2\|^2.
\end{align}

To prove (b), we use the method of induction. When $l=0$, $x_0^{(0)}(u,\lambda)= 0$ by definition. When $l=1$, $x_0^{(1)}(u,\alpha,\lambda) = \sigma(b^{(1)}(\lambda))$. Suppose $x_0^{(l-1)}(u,\alpha,\lambda) = x_0^{(l-1)}(\lambda)$ for some $l>1$, then
\begin{align}
  \nonumber
  & x_0^{(l)}(u,\alpha, \lambda) \\
  =& \sigma\left( \sum_{\lambda'}\int_{\R^2}\int_{\R} x_0^{(l-1)}(u+u', \alpha+\alpha', \lambda') W_{\lambda', \lambda}^{(l)}\left(2^{-\alpha}u', \alpha'\right)2^{-2\alpha}d\alpha' du'+b^{(l)}(\lambda)\right)\\
  =& \sigma\left( \sum_{\lambda'}x_0^{(l-1)}(\lambda') \int_{\R^2}\int_{\R} W_{\lambda', \lambda}^{(l)}\left(2^{-\alpha}u', \alpha'\right)2^{-2\alpha}d\alpha' du'+b^{(l)}(\lambda)\right)\\
  =& \sigma\left( \sum_{\lambda'}x_0^{(l-1)}(\lambda') \int_{\R^2}\int_{\R} W_{\lambda', \lambda}^{(l)}\left(u', \alpha'\right)d\alpha' du'+b^{(l)}(\lambda)\right)\\
  =& x_0^{(l)}(\lambda).
\end{align}

Part (c) is an easy corollary of part (a). More specifically, for any $l>1$,
\begin{align}
  \|x_c^{(l)}\| = \|x^{(l)}-x_0^{(l)}\| = \|x^{(l)}[x^{(l-1)}]-x_0^{(l)}[x_0^{(l-1)}]\| \le \|x^{(l-1)}-x_0^{(l-1)}\| = \|x_c^{(l-1)}\|.
\end{align}
\end{proof}

\begin{lemma}
  \label{lemma:Gm_bound}
  Suppose $\varphi\in L^2(\R)$ with $\supp(\varphi_m)\subset [-1,1]$ and $\|\varphi\|_{L^2}=1$, and $x$ is a function of three variables
  \begin{align}
    \nonumber
    x: \R^2\times \R \times [M] &\to \R\\
    (u,\alpha,\lambda) & \mapsto x(u,\alpha,\lambda)
  \end{align}
with $\|x\|^2 \coloneqq \sup_\alpha \frac{1}{M}\sum_{\lambda}\int_{\R^2}|x(u,\alpha,\lambda)|^2du$. Define $G(u,\alpha,\lambda)$ as
\begin{align}
  G(u,\alpha,\lambda) \coloneqq \int_\R x(u,\alpha+\alpha', \lambda)\varphi(\alpha')d\alpha.
\end{align}
Then we have
\begin{align}
  \label{eq:lemma_gm}
  M\|G\|^2 = \sup_\alpha \sum_{\lambda}\int_{\R^2}\left| G(u,\alpha,\lambda)\right|^2du \le 2M\|x\|^2.
\end{align}
\end{lemma}
\begin{proof}[Proof of Lemma~\ref{lemma:Gm_bound}]
  Notice that, for any $\alpha$, we have
  \begin{align}
    \sum_{\lambda}\int_{\R^2}\left|G(u,\alpha,\lambda) \right|^2du
    &= \sum_{\lambda}\int_{\R^2}\left|\int_{-1}^1x(u,\alpha+\alpha', \lambda)\varphi(\alpha')d\alpha' \right|^2 du\\
    & \le \sum_{\lambda}\int_{\R^2} \left(\int_{-1}^1\left| x(u,\alpha+\alpha',\lambda)\right|^2d\alpha' \right) \| \varphi\|_{L^2}^2 du\\
    & = \int_{-1}^1\left(\sum_{\lambda} \int_{\R^2}\left| x(u,\alpha+\alpha',\lambda)\right|^2du \right) d\alpha'\\
    & \le \int_{-1}^1 M\|x\|^2 d\alpha'= 2 M\|x\|^2.
  \end{align}
Thus
\begin{align}
  \sup_\alpha \sum_{\lambda}\int_{\R^2}\left| G(u,\alpha,\lambda)\right|^2du \le 2M\|x\|^2.
\end{align}
\end{proof}

\subsection{Proof of Theorem~\ref{thm:regularity_final}}

To prove Theorem~\ref{thm:regularity_final}, we need the following two Propositions.
\begin{prop}
  \label{prop:regularity_grad_tau}
In an ScDCFNet satisfying (A1) and (A3), we have
\begin{enumerate}[label=(\alph*)]
\item For any $l\ge 1$,
  \begin{align}
    \label{eq:regularity_grad_tau_1level}
    \left\|x^{(l)}[D_\tau x^{(l-1)}] - D_\tau x^{(l)}[x^{(l-1)}]\right\| \le 4(B_l+C_l) |\nabla \tau|_{\infty} \|x_c^{(l-1)}\|.
  \end{align}
\item For any $l\ge 1$, we have
  \begin{align}
    \label{eq:isometry}
     \|T_{\beta, v} x^{(l)}\| = 2^{\beta}\|x^{(l)}\|,
  \end{align}
  and
  \begin{align}
    \label{eq:regularity_grad_tau_group_1level}
    \left\|x^{(l)}[T_{\beta,v}\circ D_\tau x^{(l-1)}] - T_{\beta,v}D_\tau x^{(l)}[x^{(l-1)}]\right\| \le 2^{\beta+2}(B_l+C_l) |\nabla\tau|_{\infty}\|x_c^{(l-1)}\|,
  \end{align}
where the first $T_{\beta, v}$ in \eqref{eq:regularity_grad_tau_group_1level} is replaced by $D_{\beta, v}$ when $l=1$.
\item If (A2) also holds true, then
\begin{align}
  \label{eq:regularity_grad_tau_group_all}
  \left\|x^{(l)}[D_{\beta,v}\circ D_\tau x^{(0)}] - T_{\beta,v}D_\tau x^{(l)}[x^{(0)}]\right\| \le 2^{\beta+3}l |\nabla\tau|_{\infty}\|x^{(0)}\|, \quad \forall l\ge 1.
\end{align}
\end{enumerate}
\end{prop}

\begin{prop}
  \label{prop:regularity_tau}
  In an ScDCFNet satisfying (A1) and (A3), we have, for any $l\ge 1$,
  \begin{align}
    \label{eq:regularity_tau_group}
    \left\|T_{\beta,v}D_\tau x^{(l)} - T_{\beta, v}x^{(l)}  \right\| \le 2^{\beta+1}|\tau|_{\infty}D_l \|x_c^{(l-1)}\| \le 2^{\beta+1}|\tau|_{\infty}D_l \|x^{(0)}\|.
  \end{align}
If (A2) also holds true, then
 \begin{align}
    \label{eq:regularity_tau_group_all}
    \left\|T_{\beta,v}D_\tau x^{(l)} - T_{\beta, v}x^{(l)}  \right\| \le 2^{\beta+1-j_l}|\tau|_{\infty} \|x^{(0)}\|.
  \end{align}
\end{prop}

\begin{proof}[Proof of Theorem~\ref{thm:regularity_final}]
  Putting together \eqref{eq:regularity_grad_tau_group_all} and \eqref{eq:regularity_tau_group_all}, we have
  \begin{align}
    \nonumber
    &\left\| x^{(L)}[D_{\beta, v}\circ D_\tau x^{(0)}]-T_{\beta, v}x^{(L)}[x^{(0)}] \right\|\\
    \le & \left\|x^{(L)}[D_{\beta,v}\circ D_\tau x^{(0)}] - T_{\beta,v}D_\tau x^{(L)}[x^{(0)}]\right\| + \left\|T_{\beta,v}D_\tau x^{(L)}[x^{(0)}] - T_{\beta, v}x^{(L)}[x^{(0)}]  \right\|\\
    \le & 2^{\beta+3}L |\nabla\tau|_{\infty}\|x^{(0)}\| + 2^{\beta+1-j_L}|\tau|_{\infty} \|x^{(0)}\|\\
    =&2^{\beta+1}\left(4L|\nabla \tau|_\infty + 2^{-j_L}|\tau|_\infty  \right)\|x^{(0)}\|
  \end{align}
This concludes the proof of Theorem~\ref{thm:regularity_final}.
\end{proof}

Finally, we need to prove Proposition~\ref{prop:regularity_grad_tau} and Proposition~\ref{prop:regularity_tau}, where the following lemma from \cite{dcfnet} is useful.

\begin{lemma}[Lemma A.1 of \cite{dcfnet}] Suppose that $|\nabla \tau|_\infty < 1/5, \rho(u)=u-\tau(u)$, then at every point $u\in\R^2$,
  \begin{align}
    \label{eq:bd_jrho-1}
    \left||J\rho|-1\right| \le |\nabla\tau|_{\infty}(2+|\nabla \tau|_{\infty}),
  \end{align}
  where $J\rho$ is the Jacobian of $\rho$, and $|J\rho|$ is the Jacobian determinant. As a result,
  \begin{align}
    \label{eq:bd_jrho_inv-1}
    \left||J\rho|-1\right|, \left||J\rho^{-1}|-1\right| \le 4|\nabla\tau|_\infty,
  \end{align}
and,
\begin{align}
  \label{eq:bd_jrho_jrho_inv}
  \left|J\rho\right|, \left|J\rho^{-1}\right|\le 2.
\end{align}
\end{lemma}

\begin{proof}[Proof of Proposition~\ref{prop:regularity_grad_tau}]
  Just like Proposition~\ref{prop:nonexpansive}(a), the proof of Proposition~\ref{prop:regularity_grad_tau}(a) for the case $l=1$ is similar to Lemma 3.2 of \cite{dcfnet} after the change of variable \eqref{eq:change-of-variable}. We thus focus only on the proof for the case $l>1$. To simplify the notation, we denote $x^{(l)}[x^{(l-1)}]$ as $y[x]$, and replace $x_c^{(l-1)}$, $W^{(l)}$, $b^{(l)}$, $M_{l-1}$, and $M_l$, respectively, by $x_c$, $W$, $b$, $M'$, and $M$. By the definition of the deformation $D_\tau$ \eqref{eq:spatial-deformation}, we have
  \begin{align}
    &D_\tau y[x](u,\alpha,\lambda) = \sigma\left( \sum_{\lambda'}\int_{\R^2}\int_{\R} x(\rho(u)+u', \alpha+\alpha', \lambda') W_{\lambda', \lambda}\left(2^{-\alpha}u', \alpha'\right)2^{-2\alpha}d\alpha' du'+b(\lambda)\right),\\
    &y[D_\tau x](u,\alpha,\lambda) = \sigma\left( \sum_{\lambda'}\int_{\R^2}\int_{\R} x(\rho(u+u'), \alpha+\alpha', \lambda') W_{\lambda', \lambda}\left(2^{-\alpha}u', \alpha'\right)2^{-2\alpha}d\alpha' du'+b(\lambda)\right).
  \end{align}
  Thus
  \begin{align}
    &\left| (D_\tau y[x] - y[D_\tau x])(u,\alpha,\lambda)\right|^2\\\nonumber
    = &\left|\sigma\left( \sum_{\lambda'}\int_{\R^2}\int_{\R} x(\rho(u)+u', \alpha+\alpha', \lambda')W_{\lambda', \lambda}\left(2^{-\alpha}u', \alpha'\right)2^{-2\alpha}d\alpha' du'+b(\lambda)\right)\right.\\
    & - \left.\sigma\left( \sum_{\lambda'}\int_{\R^2}\int_{\R} x(\rho(u+u'), \alpha+\alpha', \lambda')W_{\lambda', \lambda}\left(2^{-\alpha}u', \alpha'\right) 2^{-2\alpha}d\alpha' du'+b(\lambda)\right)\right|^2\\\nonumber
    \le & \left| \sum_{\lambda'}\int_{\R^2}\int_{\R} \left(x(\rho(u)+u', \alpha+\alpha', \lambda')- x(\rho(u)+u', \alpha+\alpha', \lambda') \right)\right.\cdot\\
    & \left.\quad\quad\quad\quad\quad\quad\quad W_{\lambda', \lambda}\left(2^{-\alpha}u', \alpha'\right)2^{-2\alpha}d\alpha' du' \right|^2\\\nonumber
    = & \left| \sum_{\lambda'}\int_{\R^2}\int_{\R} \left(x_c(\rho(u)+u', \alpha+\alpha', \lambda')- x_c(\rho(u)+u', \alpha+\alpha', \lambda') \right)\right.\cdot\\
    & \left.\quad\quad\quad\quad\quad\quad\quad W_{\lambda', \lambda}\left(2^{-\alpha}u', \alpha'\right)2^{-2\alpha}d\alpha' du' \right|^2\\\nonumber
    = & \left| \sum_{\lambda'}\sum_m\int_{\R^2}\int_{\R} \left(x_c(\rho(u)+u', \alpha+\alpha', \lambda')- x_c(\rho(u)+u', \alpha+\alpha', \lambda') \right)\cdot\right.\\
    &\quad \quad \left.W_{\lambda', \lambda,m}\left(2^{-\alpha}u'\right) \varphi_m\left(\alpha'\right)2^{-2\alpha}d\alpha' du' \right|^2,
  \end{align}
where the second equality results from the fact that $x(u, \alpha,\lambda) - x_c(u, \alpha,\lambda) = x_0(\lambda)$ depends only on $\lambda$ (Proposition~\ref{prop:nonexpansive}(b).) Just like the proof of Proposition~\ref{prop:nonexpansive}(a), we take the integral of  $\alpha'$ first, and define
\begin{align}
  \label{eq:Gm_def_2}
  G_m(u,\alpha,\lambda') \coloneqq \int_\R x_c(u,\alpha+\alpha', \lambda')\varphi_m(\alpha')d\alpha'.
\end{align}
Thus
\begin{align}
  &\left| (D_\tau y[x] - y[D_\tau x])(u,\alpha,\lambda)\right|^2\\\nonumber
  \le & \left|\sum_{\lambda'}\sum_m \int_{\R^2}\left(G_m(\rho(u)+u',\alpha, \lambda')-G_m(\rho(u+u'),\alpha, \lambda') \right)\right.\cdot\\
  & \left.\quad\quad\quad\quad\quad\quad\quad W_{\lambda', \lambda,m}\left(2^{-\alpha}u'\right)2^{-2\alpha}du'\right|^2\\\nonumber
  = & \left|\sum_{\lambda'}\sum_m \int_{\R^2}G_m(v,\alpha, \lambda')W_{\lambda', \lambda,m}\left(2^{-\alpha}(v-\rho(u))\right)2^{-2\alpha}dv\right.\\
  &\left. - \sum_{\lambda'}\sum_m \int_{\R^2}G_m(v,\alpha, \lambda')W_{\lambda', \lambda,m}\left(2^{-\alpha}(\rho^{-1}(v)-u)\right)2^{-2\alpha}|J\rho^{-1}(v)|  dv  \right|^2\\
  = & \left| E_1(u,\alpha,\lambda)+ E_2(u,\alpha,\lambda)\right|^2,
\end{align}
where
\begin{align}\nonumber
  &E_1(u,\alpha,\lambda)= \sum_{\lambda'}\sum_m\int_{\R^2}\left[W_{\lambda',\lambda,m}\left(2^{-\alpha}(v-\rho(u)) \right) - W_{\lambda',\lambda,m}\left( 2^{-\alpha}(\rho^{-1}(v)-u)\right) \right]\cdot\\
  &\quad\quad\quad\quad\quad\quad\quad\quad\quad\quad\quad 2^{-2\alpha}G_m(v,\alpha,\lambda')dv,\\ \nonumber
  &E_2(u,\alpha,\lambda)= \sum_{\lambda'}\sum_m\int_{\R^2}W_{\lambda',\lambda,m}\left(2^{-\alpha}(\rho^{-1}(v)-u)\right) \left[1-\left| J\rho^{-1}(v)\right| \right]\cdot\\
  &\quad\quad\quad\quad\quad\quad\quad\quad\quad\quad\quad 2^{-2\alpha}G_m(v,\alpha,\lambda')dv.
\end{align}
Therefore
\begin{align}
  M\left\|D_\tau y[x] - y[D_\tau x]\right\|^2 = &\sup_\alpha \sum_{\lambda}\int_{\R^2}\left| (D_\tau y[x] - y[D_\tau x])(u,\alpha,\lambda)\right|^2du\\
  \le & \sup_\alpha \sum_{\lambda}\int_{\R^2}\left| E_1(u,\alpha,\lambda) + E_2(u,\alpha,\lambda)\right|^2du\\
  = & M\|E_1+E_2\|^2
\end{align}
Hence
\begin{align}
  \label{eq:D_less_E1_E2}
  \left\|D_\tau y[x] - y[D_\tau x]\right\| \le \|E_1+E_2\|.
\end{align}
We thus seek to estimate $\|E_1\|$ and $\|E_2\|$ individually.

To bound $\|E_2\|$, we let
\begin{align}
  k_{\lambda',\lambda,m}^{(2)}(v,u,\alpha)\coloneqq W_{\lambda',\lambda,m}\left( 2^{-\alpha}(\rho^{-1}(v)-u)\right)\left[ 1-|J\rho^{-1}(v)| \right]2^{-2\alpha}. 
\end{align}
Then
\begin{align}
  E_2(u,\alpha,\lambda) = \sum_{\lambda'}\sum_m\int_{\R^2}G_m(v,\alpha,\lambda')k_{\lambda',\lambda,m}^{(2)}(v,u,\alpha)dv,
\end{align}
and, for any given $v$ and $\alpha$
\begin{align}
  \int_{\R^2}\left|k^{(2)}_{\lambda',\lambda,m}(v,u,\alpha) \right|du =
  &\int_{\R^2}\left|W_{\lambda',\lambda,m}\left( 2^{-\alpha}(\rho^{-1}(v)-u)\right)\right|\left| 1-|J\rho^{-1}(v)| \right| 2^{-2\alpha}du\\
  = & \left| 1-|J\rho^{-1}(v)| \right| \int_{\R^2}\left| W_{\lambda',\lambda,m}(\tilde{u})\right|d\tilde{u}\\
  \le & 4|\nabla \tau|_{\infty} B_{\lambda',\lambda,m}^{(l)},
\end{align}
where the last inequality comes from \eqref{eq:bd_jrho_inv-1}. Moreover, for any given $u$ and $\alpha$,
\begin{align}
  \int_{\R^2}\left|k^{(2)}_{\lambda',\lambda,m}(v,u,\alpha) \right|dv =
  &\int_{\R^2}\left|W_{\lambda',\lambda,m}\left( 2^{-\alpha}(\rho^{-1}(v)-u)\right)\right|\left| 1-|J\rho^{-1}(v)| \right| 2^{-2\alpha}dv\\
  = & \int_{\R^2}\left| W_{\lambda',\lambda,m}(\tilde{v}-2^{-\alpha}u) \right|\cdot \left| |J\rho(2^{\alpha}\tilde{v})|-1\right|d\tilde{v}\\
  \le & 4|\nabla \tau|_{\infty} B_{\lambda',\lambda,m}^{(l)},
\end{align}
where the last inequality is again because of \eqref{eq:bd_jrho_inv-1}. Thus, for any given $\alpha$,
\begin{align}
  &\sum_{\lambda}\int_{\R^2}\left| E_2(u,\alpha,\lambda) \right|^2  du = 
  \sum_{\lambda}\int_{\R^2} \left|\sum_{\lambda'}\sum_m\int_{\R^2}G_m(v,\alpha,\lambda')k_{\lambda',\lambda,m}^{(2)}(v,u,\alpha)dv  \right|^2  du\\\nonumber
  \le & \sum_{\lambda}\int_{\R^2}\left( \sum_{\lambda'}\sum_m \int_{\R^2}\left|G_m(v,\alpha,\lambda)  \right|^2\left| k_{\lambda',\lambda,m}^{(2)}(v,u,\alpha) \right| dv\right)\cdot\\
  &\quad\quad\quad\quad \left( \sum_{\lambda'}\sum_{m}\int_{\R^2} \left| k_{\lambda',\lambda,m}^{(2)}(v,u,\alpha) \right|  dv\right)du\\\nonumber
  \le & \sum_{\lambda}\int_{\R^2}\left( \sum_{\lambda'}\sum_m \int_{\R^2}\left|G_m(v,\alpha,\lambda)  \right|^2\left| k_{\lambda',\lambda,m}^{(2)}(v,u,\alpha) \right| dv\right)\cdot\\
  &\quad\quad\quad\quad \left( \sum_{\lambda'}\sum_{m}4|\nabla \tau|_{\infty}B_{\lambda',\lambda,m}^{(l)}\right)du\\ 
  \le & 4|\nabla \tau|_{\infty}B_l\sum_m \sum_{\lambda'} \int_{\R^2}\left|G_m(v,\alpha,\lambda)  \right|^2 \left(\sum_{\lambda}\int_{\R^2} \left| k_{\lambda',\lambda,m}^{(2)}(v,u,\alpha) \right|  du \right)  dv\\
  \le & 4|\nabla \tau|_{\infty}B_l \sum_m \sum_{\lambda'} \int_{\R^2}\left|G_m(v,\alpha,\lambda)  \right|^2 \left(\sum_{\lambda} 4|\nabla \tau|_\infty B_{\lambda',\lambda,m}^{(l)} \right)  dv\\
  \le & 16 |\nabla \tau|_\infty^2 B_l \sum_m \left(\sum_{\lambda'} \int_{\R^2}\left|G_m(v,\alpha,\lambda)  \right|^2 dv\right)B_{l,m}\\
  \le & 16 |\nabla \tau|_\infty^2 B_l \sum_m M'\|G_m\|^2B_{l,m}
\end{align}
Since $\|G_m\|^2 \le 2 \|x_c \|^2$ (by Lemma~\ref{lemma:Gm_bound}), and $\sum_{m}B_{l,m}\le \frac{M}{2M'}B_l$ by definition \eqref{eq:BCDl}, we thus have, for any $\alpha$,
\begin{align}
  \label{eq:E_2_bound1}
  \sum_{\lambda}\int_{\R^2}\left| E_2(u,\alpha,\lambda) \right|^2  du
  \le 16 |\nabla \tau|_\infty^2 B_l \frac{M}{2M'}B_l \cdot 2M'\|x_c\|^2 = M(4|\nabla \tau|_{\infty}B_l\|x_c\|)^2.
\end{align}
Taking $\sup_\alpha$ on both sides gives us
\begin{align}
  \label{eq:E_2_bound}
  \|E_2\| \le 4|\nabla \tau|_\infty B_l\|x_c\|.
\end{align}

Similarly, to bound $\|E_1\|$, we introduce
\begin{align}
  k_{\lambda',\lambda,m}^{(1)}(v,u,\alpha)\coloneqq \left[W_{\lambda',\lambda,m}\left(2^{-\alpha}(v-\rho(u)) \right) - W_{\lambda',\lambda,m}\left( 2^{-\alpha}(\rho^{-1}(v)-u)\right) \right]2^{-2\alpha}.
\end{align}
Then
\begin{align}
  E_1(u,\alpha,\lambda) = \sum_{\lambda'}\sum_m\int_{\R^2}G_m(v,\alpha,\lambda')k_{\lambda',\lambda,m}^{(1)}(v,u,\alpha)dv,
\end{align}
and, for any given $v$ and $\alpha$, we have
\begin{align}
  \label{eq:k1_bound}
  \int_{\R^2}\left|k^{(1)}_{\lambda',\lambda,m}(v,u,\alpha) \right|du, ~\int_{\R^2}\left|k^{(1)}_{\lambda',\lambda,m}(v,u,\alpha) \right|dv  \le  4|\nabla \tau|_{\infty} C_{\lambda',\lambda,m}^{(l)}.
\end{align}
The proof of \eqref{eq:k1_bound} is exactly the same as that of Lemma 3.2 in \cite{dcfnet} after a change of variable, and we thus omit the detail. Similar to the procedure we take to bound $\|E_2\|$, \eqref{eq:k1_bound} leads to
\begin{align}
  \label{eq:E_1_bound}
  \|E_1\| \le 4|\nabla \tau|_\infty C_l\|x_c\|.
\end{align}
Putting together \eqref{eq:D_less_E1_E2}, \eqref{eq:E_2_bound}, and \eqref{eq:E_1_bound}, we thus have
\begin{align}
   \left\|D_\tau y[x] - y[D_\tau x]\right\| \le \|E_1+E_2\| \le \|E_1\|+\|E_2\| \le 4(B_l+C_l)|\nabla \tau|_\infty \|x_c\|.
\end{align}
This concludes the proof of (a).

To prove (b), given any $\beta\in\R$, and $v\in\R^2$, we have
\begin{align}
  \|T_{\beta,v}x^{(l)}\|^2 &= \sup_\alpha \frac{1}{M_l}\sum_{\lambda}\int_{\R^2}\left|T_{\beta,v}x^{(l)}(u,\alpha,\lambda) \right|^2 du\\
  & = \sup_\alpha \frac{1}{M_l}\sum_{\lambda}\int_{\R^2}\left|x^{(l)}(2^{-\beta}(u-v),\alpha-\beta,\lambda) \right|^2 du\\
  & = \sup_\alpha \frac{1}{M_l}\sum_{\lambda}\int_{\R^2}\left|x^{(l)}(\tilde{u},\alpha-\beta,\lambda) \right|^22^{2\beta} d\tilde{u}\\
  & = 2^{2\beta}\|x^{(l)}\|^2
\end{align}
Thus \eqref{eq:isometry} holds true. As for \eqref{eq:regularity_grad_tau_group_1level}, we have
\begin{align}
  \nonumber
  & \left\|x^{(l)}[T_{\beta,v}\circ D_\tau x^{(l-1)}] - T_{\beta,v}D_\tau x^{(l)}[x^{(l-1)}]\right\|\\
  = &\left\|T_{\beta,v} x^{(l)}[D_\tau x^{(l-1)}] - T_{\beta,v}D_\tau x^{(l)}[x^{(l-1)}]\right\|\\
  = & 2^{\beta}\left\|x^{(l)}[D_\tau x^{(l-1)}] - D_\tau x^{(l)}[x^{(l-1)}]\right\|\\
  \le & 2^{\beta+2}(B_l+C_l)|\nabla \tau|_\infty \|x_c^{(l-1)}\|,
\end{align}
where the first equality holds valid because of Theorem~\ref{thm:equivariance}, and the second equality comes from \eqref{eq:isometry}.

To prove (c), for any $0\le j\le l$, define $y_j$ as
\begin{align}
  y_j = x^{(l)}\circ x^{(l-1)}\circ \cdots \circ T_{\beta, v}\circ D_\tau x^{(j)}\circ \cdots \circ x^{(0)}.
\end{align}
We thus have
\begin{align}
  &\left\|x^{(l)}[D_{\beta,v}\circ D_\tau x^{(0)}] - T_{\beta,v}D_\tau x^{(l)}[x^{(0)}]\right\| = \|y_l - y_0\|  \le \sum_{j=1}^l\|y_j - y_{j-1}\|\\
  = & \sum_{j=1}^l \left\|x^{(l)}\circ \cdots \circ T_{\beta, v}\circ D_\tau x^{(j)}\circ\cdots \circ x^{(0)} - x^{(l)}\circ \cdots \circ x^{(j)}\circ T_{\beta, v}\circ D_\tau x^{(j-1)}\circ \cdots \circ x^{(0)}\right\|\\
  \le &\sum_{j=1}^l \left\|T_{\beta, v}\circ D_\tau x^{(j)}[x^{(j-1)}] -  x^{(j)}[T_{\beta, v}\circ D_\tau x^{(j-1)}]\right\|\\
  \le &\sum_{j=1}^l 2^{\beta+2}(B_j+C_j)|\nabla \tau|_\infty\|x_c^{(j-1)}\|\\
  \le & \sum_{k=1}^l 2^{\beta+2} \cdot 2 |\nabla \tau|_\infty \|x^{(0)}\| = 2^{\beta+3}l |\nabla \tau|_\infty \|x^{(0)}\|,
\end{align}
where the second inequality is because of Proposition~\ref{prop:nonexpansive}(a), the third inequality is due to \eqref{eq:regularity_grad_tau_group_1level}, and the last inequality holds true because $B_l, C_l \le A_l \le 1$ under (A2) (Lemma~\ref{lemma:Al-bounds-everything}.) This concludes the proof of Proposition~\ref{prop:regularity_grad_tau}.
\end{proof}

\begin{proof}[Proof of Proposition~\ref{prop:regularity_tau}]
  The second inequality in \eqref{eq:regularity_tau_group} is due to Proposition~\ref{prop:nonexpansive}(c). Because of \eqref{eq:isometry}, the first inequality in \eqref{eq:regularity_tau_group} is equivalent to
  \begin{align}
    \label{eq:regularity_tau}
    \left\|D_\tau x^{(l)} - x^{(l)}  \right\| \le 2|\tau|_{\infty}D_l \|x_c^{(l-1)}\|
  \end{align}
Just like Proposition~\ref{prop:regularity_grad_tau}(a), the proof of \eqref{eq:regularity_tau} for the case $l=1$ is similar to Proposition 3.4 of \cite{dcfnet} after the change of variable \eqref{eq:change-of-variable}. A similar strategy as that of Proposition~\ref{prop:regularity_grad_tau}(a) can be used to extend the proof to the case $l>1$. More specifically, denote $x^{(l-1)}, x_c^{(l-1)}, W^{(l)}, b^{(l)}$, respectively, as $x, x_c, W,$ and  $b$ to simplify the notation. We have
\begin{align}
  \nonumber
  & \left| \left(D_\tau x^{(l)}[x] - x^{(l)}[x] \right)(u,\alpha,\lambda) \right|^2\\
  \nonumber
  = & \left| \sigma \left(\sum_{\lambda'}\int_{\R^2}\int_\R x\left(\rho(u)+u',\alpha+\alpha',\lambda'\right)W_{\lambda',\lambda}\left( 2^{-\alpha}u',\alpha'\right) 2^{-2\alpha}du'd\alpha'+b(\lambda)\right)\right.\\
  & - \left.\sigma \left(\sum_{\lambda'}\int_{\R^2}\int_\R x\left(u+u',\alpha+\alpha',\lambda'\right)W_{\lambda',\lambda}\left( 2^{-\alpha}u',\alpha'\right) 2^{-2\alpha}du'd\alpha'+b(\lambda)\right)\right|\\ \nonumber
  \le & \left| \sum_{\lambda'}\int_{\R^2}\int_\R x\left(\rho(u)+u',\alpha+\alpha',\lambda'\right)W_{\lambda',\lambda}\left( 2^{-\alpha}u',\alpha'\right) 2^{-2\alpha}du'd\alpha'\right.\\
  & - \left.\sum_{\lambda'}\int_{\R^2}\int_\R x\left(u+u',\alpha+\alpha',\lambda'\right)W_{\lambda',\lambda}\left( 2^{-\alpha}u',\alpha'\right) 2^{-2\alpha}du'd\alpha'\right|\\ \nonumber
  = & \left| \sum_{\lambda'}\int_{\R^2}\int_\R x_c\left(\rho(u)+u',\alpha+\alpha',\lambda'\right)\sum_mW_{\lambda',\lambda,m}\left( 2^{-\alpha}u'\right)\varphi_m\left(\alpha'\right) 2^{-2\alpha}du'd\alpha'\right.\\
  & - \left.\sum_{\lambda'}\int_{\R^2}\int_\R x_c\left(u+u',\alpha+\alpha',\lambda'\right)\sum_mW_{\lambda',\lambda,m}\left( 2^{-\alpha}u'\right)\varphi_m\left(\alpha'\right) 2^{-2\alpha}du'd\alpha'\right|\\
  = & \left| \sum_{\lambda'}\sum_m \int_{\R^2}G_m(v,\alpha,\lambda')k_{\lambda',\lambda,m}(v,u,\alpha) du'\right|,
\end{align}
where
\begin{align}
  G_m(u,\alpha,\lambda') &\coloneqq \int_\R x_c(u,\alpha+\alpha', \lambda')\varphi_m(\alpha')d\alpha',\\
  k_{\lambda',\lambda,m}(v,u,\alpha) & \coloneqq  2^{-2\alpha} \left[ W_{\lambda',\lambda,m}\left( 2^{-\alpha}(v-\rho(u)) \right) -W_{\lambda',\lambda,m}\left( 2^{-\alpha}(v-u) \right)\right].
\end{align}
Similar to \eqref{eq:k1_bound}, we have the following bound
\begin{align}
  \label{eq:k_bound}
  \int_{\R^2}\left|k_{\lambda',\lambda,m}(v,u,\alpha) \right|du, ~\int_{\R^2}\left|k_{\lambda',\lambda,m}(v,u,\alpha) \right|dv  \le  2|\nabla \tau|_{\infty} D_{\lambda',\lambda,m}^{(l)}.
\end{align}
Again, the proof of \eqref{eq:k_bound} is the same as that of Proposition 3.4 in \cite{dcfnet} after a change of variable. The rest of the proof follows from a similar argument as in \eqref{eq:E_2_bound1} and \eqref{eq:E_2_bound}.
\end{proof}

\subsection{Proof of Theorem~\ref{thm:regularity_final_truncation}}

  Before proving Theorem~\ref{thm:regularity_final_truncation}, we need the following lemma.
  \begin{lemma}
    \label{lemma:truncation_error}
    Under the same assumption of Theorem~\ref{thm:regularity_final_truncation}, we have, for any $l$,
    \begin{align}
      \label{eq:o2t}
      \left\|\tilde{x}^{(l)}[x^{(0)}] - x^{(l)}[x^{(0)}]\right\|^2 = \sup_{\alpha \le T}\frac{1}{M_l}\sum_{\lambda = 1}^{M_l}\int\left|\tilde{x}^{(l)}(u,\alpha, \lambda) - x^{(l)}(u,\alpha, \lambda) \right|^2du = O(2^{-T}),
    \end{align}
    where we  slightly abuse the notation $\tilde{x}^{(l)}$ and $x^{(l)}$ to denote the $l$-th layer outputs given the input $x^{(0)}$ in the first equality.
  \end{lemma}
  \begin{proof}[Proof of Lemma~\ref{lemma:truncation_error}]
    We prove this lemma by induction. When $l=1$, we have, for any $\alpha\in [-T, T]$,
    \begin{align}
      \label{eq:exact_truncation}
      \tilde{x}^{(1)}(u,\alpha,\lambda) = x^{(1)}(u,\alpha,\lambda).
    \end{align}
    When $\alpha \le -T$, we have
    \begin{align}
      & \left|\tilde{x}^{(1)}(u,\alpha, \lambda) - x^{(1)}(u,\alpha,\lambda)\right| = \left| x^{(1)}(u,-T,\lambda) - x^{(1)}(u,\alpha,\lambda)\right|\\ \nonumber
      = & \left|\sigma \left(\sum_{\lambda'}\int_{\R^2}2^{2T}x^{(0)}(u+u', \lambda')W_{\lambda', \lambda}(2^Tu')du' + b(\lambda)\right)\right.\\
      & \left. -\sigma \left(\sum_{\lambda'}\int_{\R^2}2^{-2\alpha}x^{(0)}(u+u', \lambda')W_{\lambda', \lambda}(2^{-2\alpha}u')du' + b(\lambda)\right)\right|\\
      \le & \left| \sum_{\lambda'} x^{(0)}(\cdot, \lambda')* W_{\lambda',\lambda, -T}(u) - x^{(0)}(\cdot, \lambda')* W_{\lambda',\lambda, \alpha}(u) \right|,
    \end{align}
    where $W_{\lambda',\lambda, \alpha}(u) = 2^{-2\alpha}W_{\lambda',\lambda}(2^{-\alpha}u)$ forms a mollifier in $\R^2$. Thus, we have
  \begin{align}\nonumber
    & \int_{\R^2}\left|\tilde{x}^{(1)}(u,\alpha, \lambda) - x^{(1)}(u,\alpha,\lambda)\right|^2du\\
    \le & \int_{\R^2}\left| \sum_{\lambda'} x^{(0)}(\cdot, \lambda')* W_{\lambda',\lambda, -T}(u) - x^{(0)}(\cdot, \lambda')* W_{\lambda',\lambda, \alpha}(u) \right|^2 du\\
    \le & \int_{\R^2} \left(\sum_{\lambda'} \left|x^{(0)}(\cdot, \lambda')* W_{\lambda',\lambda, -T}(u) - x^{(0)}(\cdot, \lambda')* W_{\lambda',\lambda, \alpha}(u) \right|^2\right) M' du\\
    = & M'\sum_{\lambda'}\left\|x^{(0)}(\cdot, \lambda')* W_{\lambda',\lambda, -T} - x^{(0)}(\cdot, \lambda')* W_{\lambda',\lambda, \alpha} \right\|^2_{2}\\\nonumber   
    \le & M'\sum_{\lambda'}\left(\left\|x^{(0)}(\cdot, \lambda')* W_{\lambda',\lambda, -T} - A_{\lambda',\lambda}x^{(0)}(\cdot, \lambda')\right\|_{2} \right.\\
    & \quad\quad\quad\quad\quad  + \left.\left\|x^{(0)}(\cdot, \lambda')* W_{\lambda',\lambda, \alpha} - A_{\lambda',\lambda}x^{(0)}(\cdot, \lambda')\right\|_{2} \right)^2,
  \end{align}
  where $A_{\lambda',\lambda} = \int_{\R^2}W^{(1)}_{\lambda',\lambda}(u)du$. Due to the $L^2$ convergence of mollification \cite{evans10}, i.e.,
  \begin{align}
    \left\|f*W_{\lambda',\lambda, \alpha} - A_{\lambda',\lambda} f\right\|_2 \le C 2^{\alpha}\left\|f \right\|_{H^1} = O(2^{\alpha}), ~~\text{as}~\alpha\to -\infty,
  \end{align}
  we have
  \begin{align}
    \label{eq:o2t_step1}
    \sup_{\alpha\le -T}\int_{\R^2}\left|\tilde{x}^{(1)}(u,\alpha, \lambda) - x^{(1)}(u,\alpha,\lambda)\right|^2du = O(2^{-T}).
  \end{align}
  Combining \eqref{eq:o2t_step1} and \eqref{eq:exact_truncation}, we have
  \begin{align}
    \|\tilde{x}^{(1)} - x^{(1)}\|^2 = \sup_{\alpha \le T}\frac{1}{M}\sum_{\lambda}\int\left|\tilde{x}^{(1)}(u,\alpha,\lambda) - x^{(1)}(u,\alpha,\lambda)\right|^2du = O(2^{-T}).
  \end{align}
  Next, we want to show that \eqref{eq:o2t} holds for $l > 1$ assuming it holds for $l-1$. Indeed, for any $\alpha\in [-T, T]$, we have
  \begin{align}
    \tilde{x}^{(l)}(u,\alpha, \lambda) - x^{(l)}(u, \alpha, \lambda) = x^{(l)}[\tilde{x}^{(l-1)}](u,\alpha, \lambda) - x^{(l)}[x^{(l-1)}](u, \alpha, \lambda).
  \end{align}
  Taking the supremum over $\alpha \in [-T, T]$, we have
  \begin{align}
    \nonumber
    & \sup_{\alpha \in [-T, T]}\frac{1}{M}\sum_{\lambda}\int \left|\tilde{x}^{(l)}(u,\alpha, \lambda) - x^{(l)}(u,\alpha, \lambda) \right|^2du\\
    = & \sup_{\alpha \in [-T, T]}\frac{1}{M}\sum_{\lambda}\int \left|x^{(l)}[\tilde{x}^{(l-1)}](u,\alpha, \lambda) - x^{(l)}[x^{(l-1)}](u,\alpha, \lambda) \right|^2du\\
    \le & \left\|x^{(l)}[\tilde{x}^{(l-1)}] - x^{(l)}[x^{(l-1)}]  \right\|^2\\ \label{eq:o2t_step2}
    \le & \left\|\tilde{x}^{(l-1)} - x^{(l-1)}  \right\|^2 = O(2^{-T}),
  \end{align}
  where the last inequality results from the non-expansiveness of $x^{(l)}$ (c.f. Proposition~\ref{prop:nonexpansive},) and the last equality comes from our assumption that \eqref{eq:o2t} holds for $l-1$. For any $\alpha \le -T$, we have
  \begin{align}
    \nonumber
    & \frac{1}{M}\sum_{\lambda}\int\left| \tilde{x}^{(l)}(u, \alpha, \lambda) - x^{(l)}(u, \alpha, \lambda) \right|^2du\\
    = &  \frac{1}{M}\sum_{\lambda}\int\left| \tilde{x}^{(l)}(u, -T, \lambda) - x^{(l)}(u, \alpha, \lambda) \right|^2du\\
    = & \frac{1}{M}\sum_{\lambda}\int\left| \tilde{x}^{(l)}(u, -T, \lambda) - x^{(l)}(u, -T, \lambda) + x^{(l)}(u, -T, \lambda) - x^{(l)}(u, \alpha, \lambda) \right|^2du\\
    \le & \frac{1}{M}\sum_{\lambda}\int\left| \tilde{x}^{(l)}(u, -T, \lambda) - x^{(l)}(u, -T, \lambda) + x^{(l)}(u, -T, \lambda) - x^{(l)}(u, \alpha, \lambda) \right|^2du\\
    \le & \frac{2}{M}\sum_{\lambda}\int\left| \tilde{x}^{(l)}(u, -T, \lambda) - x^{(l)}(u, -T, \lambda)\right|^2 + \left|x^{(l)}(u, -T, \lambda) - x^{(l)}(u, \alpha, \lambda) \right|^2du\\
     = & O(2^{-T}) + \frac{2}{M}\sum_{\lambda}\int\left|x^{(l)}(u, -T, \lambda) - x^{(l)}(u, \alpha, \lambda) \right|^2du,
  \end{align}
  where, in the last equality, the first term $O(2^{-T})$ does not depend on $\alpha\le -T$ and comes from \eqref{eq:o2t_step2}. Taking the supremum over $\alpha\le -T$, we have
  \begin{align}
    \nonumber
    & \sup_{\alpha\le -T}\frac{1}{M}\sum_{\lambda}\int\left| \tilde{x}^{(l)}(u, \alpha, \lambda) - x^{(l)}(u, \alpha, \lambda) \right|^2du\\
    \le & O(2^{-T}) + \sup_{\alpha\le -T}\frac{2}{M}\sum_{\lambda}\int\left|x^{(l)}(u, -T, \lambda) - x^{(l)}(u, \alpha, \lambda) \right|^2du\\ \label{eq:o2t_step3}
    = & O(2^{-T}),
  \end{align}
  where the last equality holds for the same reason as \eqref{eq:o2t_step1}. Combining \eqref{eq:o2t_step2} and \eqref{eq:o2t_step3}, we thus have
  \begin{align}
    \left\|\tilde{x}^{(l)}[x^{(0)}] - x^{(l)}[x^{(0)}]\right\|^2 = \sup_{\alpha \le T}\frac{1}{M_l}\sum_{\lambda = 1}^{M_l}\int\left|\tilde{x}^{(l)}(u,\alpha, \lambda) - x^{(l)}(u,\alpha, \lambda) \right|^2du = O(2^{-T}).   
  \end{align}
\end{proof}
With Lemma~\ref{lemma:truncation_error} proven, the proof of Theorem~\ref{thm:regularity_final_truncation} is merely an easy application of triangle inequality:
\begin{proof}[Proof of Theorem~\ref{thm:regularity_final_truncation}]
  \begin{align}
    \nonumber
    & \left\| \tilde{x}^{(L)}[D_{\beta, v}\circ D_\tau x^{(0)}]-T_{\beta, v}\tilde{x}^{(L)}[x^{(0)}] \right\| \\\nonumber
    \le & \left\| \tilde{x}^{(L)}[D_{\beta, v}\circ D_\tau x^{(0)}] - x^{(L)}[D_{\beta, v}\circ D_\tau x^{(0)}]\right\| + \left\|x^{(L)}[D_{\beta, v}\circ D_\tau x^{(0)}] - T_{\beta, v}x^{(L)}[x^{(0)}] \right\| \\
    & + \left\|T_{\beta, v}x^{(L)}[x^{(0)}] - T_{\beta, v}\tilde{x}^{(L)}[x^{(0)}] \right\| \\ \label{eq:thm3_last}
    \le & O(2^{-T})+2^{\beta+1}\left( 4L|\nabla \tau|_\infty + 2^{-j_L}|\tau|_\infty  \right)\|x^{(0)}\| + 2^\beta O(2^{-T})\\
    = & 2^{\beta+1}\left( 4L|\nabla \tau|_\infty + 2^{-j_L}|\tau|_\infty  \right)\|x^{(0)}\| + O(2^{-T}),
  \end{align}
  where the last inequality \eqref{eq:thm3_last} comes from Lemma~\ref{lemma:truncation_error} and Theorem~\ref{thm:regularity_final}.
\end{proof}

\section{Experimental Details in Section~\ref{sec:experiments} }

\subsection{Verification of $\cst$-equivariance}
\label{sec:app_ex1}
 The ScDCFNet used in this experiment has two convolutional layers, each of which is composed of a $\cst$-equivariant convolution \eqref{eq:1st-layer} or \eqref{eq:lth-layer}, a batch-normalization, and a $2\times 2$ spatial average-pooling. The expansion coefficients $a_{\lambda',\lambda}^{(1)}(k)$ and $a_{\lambda',\lambda}^{(2)}(k,m)$ are sampled from a Gaussian distribution and truncated to $K = 8$ and $K_\alpha = 3$ leading coefficients for $u$ and $\alpha$ respectively. Similarly, a regular CNN with two  convolutional layers and randomly generated $5\times 5$ convolutional kernels is used as a baseline for comparison.

\subsection{Image Reconstruction}
\label{sec:app_ex3}
The network architectures for the ScDCFNet and regular CNN auto-encoders are shown in Table~\ref{tab:architecture_auto}. The filter expansion in the ScDCFNet auto-encoder is truncated to $K=8$ and $K_\alpha=3$. SGD with decreasing learning rate from $10^{-2}$ to $10^{-4}$ is used to train both networks for 20 epochs.

\begin{table}[H]
  \centering
  \footnotesize
  \begin{tabular}{cll}
    \toprule
    Layer & Regular auto-encoder & ScDCF auto-encoder\\
    \midrule\midrule
    1 & c7x7x1x8 ~ReLU ~ap2x2 & sc(15)13x13x1x4 ~ReLU ~ap2x2\\
    \midrule
    2 & c7x7x8x16 ~ReLU ~ap2x2 & sc(15)13x13x3x4x8 ~ReLU ~ap2x2\\
    \midrule
    3 & fc128 ~ReLU ~fc4096 ~ReLU & fc128 ~ReLU ~fc4096 ~ReLU\\
    \midrule
    4 & ct7x7x16x8 ~ReLU ~us2x2 & ct7x7x16x8 ~ReLU ~us2x2\\
    \midrule
    5 & ct7x7x8x1 ~ReLU ~us2x2 & ct7x7x8x1 ~ReLU ~us2x2\\
    \bottomrule
  \end{tabular}
  \vspace{-.5em}
  \caption{\small Architectures of the auto-encoders used for the experiment in Section~\ref{sec:image_reconstruction}. The encoded representation is the output of the second layer. \textbf{cLxLxM'xM:} a regular convolutional layer with M' input channels, M output channels, and LxL spatial kernels. \textbf{apLxL:} LxL average-pooling. \textbf{sc($N_s$)LxLxM'xM:} the first-layer convolution \eqref{eq:1st-layer} in ScDCFNet, where $N_s$ is the number of scale channels, and LxL is the spatial kernel size. \textbf{sc($N_s$)LxLx$L_\alpha$xM'xM:} the $l$-th layer ($l>1$) convolution \eqref{eq:lth-layer} in ScDCFNet, where the extra symbol $L_\alpha$ stands for the filter size in $\alpha$. \textbf{fcM:} a fully connected layer with M output channels. \textbf{ctLxLxM'xM:} transposed-convolutional layers with M' input channels, M output channels, and LxL spatial kernels. \textbf{us2x2}: 2x2 spatial upsampling. Batch-normalization (not shown in the table) is used after each convolutional layer.}
  \label{tab:architecture_auto}
\end{table}

\bibliography{scdcf_jmlr}
\bibliographystyle{abbrv}
\end{document}